\documentclass[twoside,11pt]{article}

\usepackage{blindtext}

\usepackage[preprint]{jmlr2e}

\usepackage{multirow}
\usepackage{bbm}
\usepackage{graphicx}
\usepackage{amssymb}
\usepackage{amsmath} 
\usepackage{booktabs}
\usepackage{lscape}
\usepackage{pifont}
\usepackage{algorithm}
\usepackage{algpseudocode}
\usepackage{graphicx}
\usepackage{colortbl}
\usepackage{hyperref}
\usepackage{xcolor}

\newcommand{\E}[2]{\mathbb{E}_{#1}\left[#2\right]}

\DeclareMathOperator*{\argmax}{argmax}

\ShortHeadings{}{}
\firstpageno{1}

\begin{document}

\title{Preference Tuning with Human Feedback on Language, Speech, and Vision Tasks: A Survey}

\author{\name Genta Indra Winata$^{*1}$ \email genta.winata@capitalone.com \\
        \name Hanyang Zhao$^{*2}$ \email hz2684@columbia.edu \\
        \name Anirban Das$^{*1}$ \email anirban.das3@capitalone.com \\
       \name Wenpin Tang$^{2}$ \email wt2319@columbia.edu \\
       \name David D. Yao$^{2}$ \email ddy1@columbia.edu \\
       \name Shi-Xiong Zhang$^{1}$ \email shixiong.zhang@capitalone.com \\
       \name Sambit Sahu$^{1}$ \email sambit.sahu@capitalone.com \\
       \addr $^{1}$Capital One $\quad^{2}$Columbia University 
       }

\editor{Preprint}

\maketitle

\begin{abstract}
Preference tuning is a crucial process for aligning deep generative models with human preferences. This survey offers a thorough overview of recent advancements in preference tuning and the integration of human feedback. The paper is organized into three main sections: 1) \textbf{introduction and preliminaries}: an introduction to reinforcement learning frameworks, preference tuning tasks, models, and datasets across various modalities: language, speech, and vision, as well as different policy approaches, 2) \textbf{in-depth exploration of each preference tuning approach}: a detailed analysis of the methods used in preference tuning, and 3) \textbf{applications, discussion, and future directions}: an exploration of the applications of preference tuning in downstream tasks, including evaluation methods for different modalities, and an outlook on future research directions. Our objective is to present the latest methodologies in preference tuning and model alignment, enhancing the understanding of this field for researchers and practitioners. We hope to encourage further engagement and innovation in this area.
\end{abstract}

\begin{keywords}
  preference tuning, human preference, reinforcement learning, multi-modality, multilingual, large language models, vision language models, speech language models, generative models, survey, DPO, RLHF.
\end{keywords}

\tableofcontents

\section{Introduction}
Learning from human feedback is a crucial step in aligning generative models with human preferences to generate output that closely resembles human speech and writing. Despite the powerful learning capabilities of generative models in self-supervised learning, these models frequently misinterpret instructions, leading to hallucinations in generation~\citep{ji2023survey,yao2023llm}. Additionally, ensuring the safety of the generated content remains a significant challenge for these models. Extensive research on preference tuning using human feedback has demonstrated that adversarial samples can be utilized to jailbreak systems~\citep{rando2023universal,wei2024jailbroken}. Ideally, generative models need to be controlled to ensure that their outputs are safe and do not cause harm. Models often exhibit unintended behaviors, such as fabricating facts~\citep{chen2023can,sun2024exploring}, producing biased or toxic text~\citep{hartvigsen2022toxigen}, or failing to follow user instructions~\citep{ji2023towards,tonmoy2024comprehensive}. Additionally, maintaining the privacy of data is crucial to ensure the safe operation of models and protect user privacy~\citep{brown2022does}. In the text-to-image generation task, large-scale models often struggle to produce images that are well-aligned with text prompts~\citep{feng2022training}, particularly in compositional image generation~\citep{liu2022compositional,lee2023aligning}, object recognition~\citep{qiao2024robust}, and coherent generation~\citep{liu2023character}. Similarly, in text-to-speech tasks, \cite{zhang2024speechalign,chen2024enhancing} integrate subjective human evaluation into the training loop to better align synthetic speech with human preferences.

The application of preference tuning has been widely used in language tasks by training instruction-tuned large language models (LLMs), such as Llama~\citep{touvron2023llama2,dubey2024llama}, Phi~\citep{abdin2024phi}, Mistral~\citep{jiang2023mistral}, Nemotron~\citep{parmar2024nemotron,adler2024nemotron}, Gemma~\citep{team2024gemma}.
Commercial models like GPT-4~\citep{achiam2023gpt}, 
Gemini~\citep{team2023gemini,reid2024gemini}, Claude~\citep{anthropic2024claude}, Command-R, and Reka~\citep{ormazabal2024reka} have leveraged human preference alignment to enhance their performance. Alignment of LLM improves task-specific skills, coherence, fluency, and helps avoid undesired outputs. Additionally, alignment research has benefited multilingual LLMs, such as Aya~\citep{aryabumi2024aya,ustun2024aya}, BLOOMZ, and mT0~\citep{muennighoff2023crosslingual}, as well as regional LLMs like Cendol~\citep{cahyawijaya2024cendol} and SEALLM~\citep{nguyen2023seallms}. Common approaches to achieving LLM alignment involve reinforcement learning techniques that guide language models to follow preferred samples by maximizing rewards. Reinforcement Learning from Human Feedback (RLHF)~\citep{christiano2017deep} is the initial approach that is used to align models with human preference, which is further applied to the deep learning space that has been popularized by its successes in  LLMs~\citep{ouyang2022training,bai2022training} via PPO~\citep{schulman2017proximal}, REINFORCE~\citep{kool2019buy}, Online Directed Preference Optimization (online DPO)~\citep{guo2024direct}, and Supervised Fine-Tuning (SFT)-like approach~\citep{dong2023raft}. It typically involves three key aspects: human feedback collection, reward modeling, and {\it online} RL for policy optimization. Recent methods, however, allow for training the reward model alongside the policy model in an {\it offline} manner, as demonstrated by DPO~\citep{rafailov2024direct}, and jointly training with offline and online policies training~\citep{zhao2023slic}. Moreover, preference tuning is also applied to  vision-text tasks, and has been shown to improve the representation of both image and text using the alignment score of image and text embeddings~\citep{ramesh2022hierarchical,saharia2022photorealistic,yu2022scaling} measured by pre-trained vision-text models, such as CLIP~\citep{radford2021learning} and CoCa~\citep{yu2022coca}. \cite{wu2023human} utilize LoRA~\citep{hu2021lora} to align Stable Diffusion~\citep{lee2023aligning}, a vision-text pre-trained model. The application in speech has not been  much explored, and there is only a handful works in the literature.
\cite{zhang2024speechalign} focus on investigating alignment between codes and the text.

In this paper, we survey the recent advances of preference tuning with human feedback in different modalities. It provides not only a comprehensive introduction including preliminaries to get readers familiar with the topic, but also an in-depth review on the latest proposed approaches and in-depth discussions. To summarize, the paper comprises the following contributions:
\begin{itemize}
\item We provide a comprehensive overview of preference tuning for models on different modalities, such as language, speech, and vision tasks, and expand our survey to all existing preference tuning methods, including reinforcement learning (RL) approaches.
\item We formulate and taxonomize a systematic framework and classification for preference tuning for deep generative models from the existing literature.
\item We present various applications of preference tuning to improve generation aspects using human feedback. We also describe the automatic and human-based evaluations to measure the quality of generation in deep generative models.
\item We discuss the opportunities and future directions for preference tuning.
\end{itemize}
Through this survey, we aim to present the recent methodologies on preference tuning and alignment for deep generative models, enabling researchers and practitioners to better understand this topic and further innovate.

\section{Preliminaries}
This section outlines the preliminaries of preference tuning, including the formal definitions of the tasks and the notations used throughout this paper. Additionally, we provide a taxonomy for classifying preference tuning methods.
\subsection{Tasks and Definition}
In general, the entire preference tuning mechanism for generative models can be formulated as a RL problem described as follows.

\subsubsection{RL Framework Concepts}
\paragraph{Policy Model} The policy model $\pi_\theta$ is a generative model that takes in an input prompt $x$ and returns a sequence of output or probability distributions $y$. We define a generative model as a policy model $\pi_\theta$ where it is parameterized by $\theta$ with a policy model $\pi$. Given a prompt $x$, a generative model generates an output $y$ as following:
\begin{align}
    \pi_\theta(y|x) = \prod_t \pi_\theta(y_t|x,y_{<t}),
\end{align}
where $y_t$ is the $t$-th token in the response and $y_{<t}$ is tokens in the response before $y_t$. For example, for the text-based tasks, the input prompt is a text sequence $x$ and the output is a probability distribution over text vocabulary of LLM $y$; and for the vision-text-based tasks, such as text-to-image tasks, the input $x$ is the text sequence, and $y$ is the generated image.

\paragraph{Reward Model} The reward model (RM) processes both the input $x$ and the target $y$, passing them through the model to obtain a reward $r_\theta(y|x)$, which reflects the notion of preferability. This preferability score can also be interpreted as a relative score assigned to the target $y$ given the input $x$. Less preferred outcomes receive a lower score compared to more preferred samples.

\paragraph{Action Space} The action refers to all tokens corresponding to the vocabulary of generative models. For text tasks, the action space encompasses the entire vocabulary of the LLM. For vision tasks (similarly for speech tasks), the action space consists of real values representing the image, for example, the next hierarchy in diffusion generative models (if understanding diffusion models as Hierarchical Variational Autoencoders \citep{luo2022understanding}).

\paragraph{Environment} The distribution encompasses all possible input token sequences for generative models. 
In text-based tasks, these input token sequences correspond to text sequences, highly depending on the sampling methods for the inference.
In vision tasks, they correspond to possible images.

\subsubsection{Preference Data} 
In the preference tuning pipeline, we utilize the supervised data $\mathcal{D}_\text{sft}$ and the preference data $\mathcal{D}_\text{pref}$. 
We denote the supervised data $\mathcal{D}_\text{sft} = [(x^1, y^1), \cdots, (x^{M}, y^{M})]$ as a list of input and label pairs. Specifically for the text SFT data, $x$ can be represented as prompts. The prompt $x^i=(I^i,F^i,Q^i)$ consists of the concatenation of an instruction $I^i$, few-shot samples $F^i$, and a query $Q^i$. Then, we denote the preference data $\mathcal{D}_\text{pref} = [(x^1, y_w^1, y_l^1), \cdots, (x^N, y_w^N, y_l^N)]$, a list of input $x^{i}$ with preferred response $y_w^{i}$ and dispreferred response $y_l^{i}$, and they are either sampled from the reference policy model $\pi_\text{ref}$ or collected by human annotation. Generally, given the preference data, we can obtain a reward $r$ associated to the response with the input.

\subsubsection{Terminology and Notation}
Table~\ref{tab:table-of-notations} lists the common notations used in this survey paper. The table serves as a quick reference guide for understanding the mathematical expressions and technical terms used throughout the paper.
\begin{table*}[!ht]
\centering
\resizebox{0.98\textwidth}{!}{
    \begin{tabular}{lcl}
    \toprule
    \textbf{Name} & \textbf{Notation} & \multicolumn{1}{c}{\textbf{Description}} \\ \midrule
    Input Sequence & $x$ & Input sequence that is passed to the model. \\ 
    Output Sequence & $y$ & Expected label or output of the model. \\ \midrule
    Dispreferred Response & $y_l$ & Negative samples for reward model training. \\ 
    Preferred Response & $y_w$ & Positive samples for reward model training. \\ \midrule
    Optimal Policy Model & $\pi^*$ & Optimal policy model. \\
    Policy Model & $\pi_\theta$ & Generative model that takes the input prompt and \\
    & & returns a sequence of output or probability distribution. \\ 
    Reference Policy Model & $\pi_\text{ref}$ & Generative model that is used as a reference to \\
    & & ensure the policy model is not deviated significantly. \\ \midrule
    Preference Dataset & $\mathcal{D}_\text{pref}$ & Dataset with a set of preferred and dispreferred. \\
    & & responses to train a reward model.\\ 
    SFT Dataset & $\mathcal{D}_\text{sft}$ & Dataset with a set of input and label for supervised \\
    & & fine-tuning.\\ \midrule
    Loss Function & $\mathcal{L}$ & Loss function. \\ 
    Regularization Hyper-parameters & $\alpha, \beta_\text{reg}$ & Regularization Hyper-parameters for preference tuning. \\
    Reward & $r$ & Reward score. \\  
    Target Reward Margin & $\gamma$ & The margin separating the winning and losing responses. \\
    Variance & $\beta_i$ & Variance (or noise schedule) used in diffusion models. \\
    \bottomrule
    \end{tabular}
}
\caption{Table of Terminology and Notation.}
\label{tab:table-of-notations}
\end{table*}

\subsection{Taxonomy}

\begin{figure*}[!t]
    \centering
    \includegraphics[width=\linewidth]{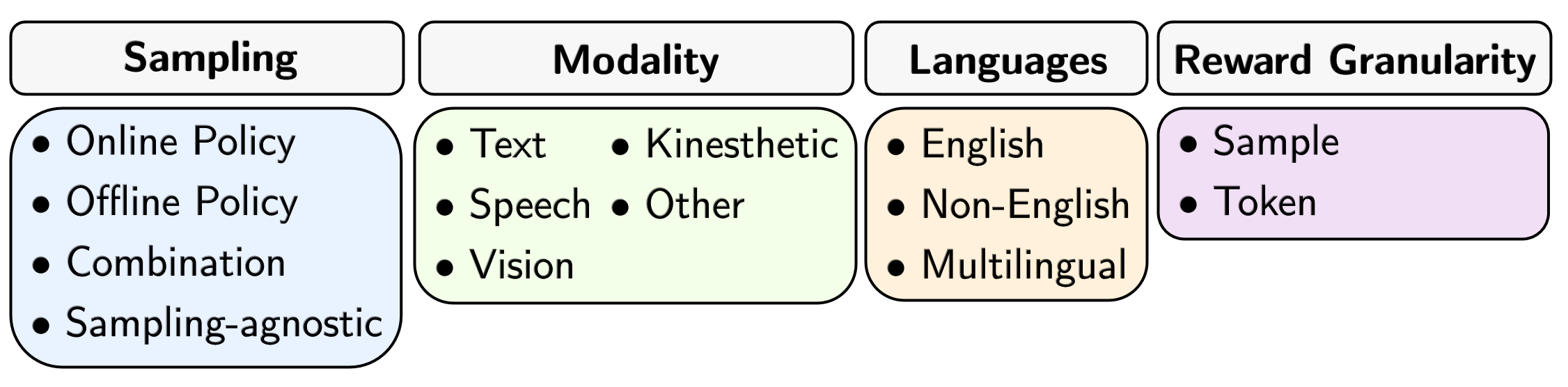} 
    \caption{Taxonomy of the Preference Tuning methods.}
    \label{fig:taxonomy}
\end{figure*}

We define the following categories for all of the preference tuning approaches as shown in Table~\ref{tab:category}. Figure~\ref{fig:taxonomy} shows the five categories we study in this survey paper and described in the following:

\paragraph{Sampling}
Likewise in the literature of RL, we categorize the methods based on how we sample the data and use them to train or obtain the reward: {\it offline} and {\it online} human alignments. The categorization is related to how we compute the reward and use it in the policy models. In online human alignment setting, the agent that collects a batch of examples by interacting with the environment and uses them to update the policy. The reward of the examples can be collected by the reward model or samples generated by the policy model. While for the offline human alignment setting, the data are collected from offline human demonstrations. For online methods, we also categorize the methods as either {\it on-policy} when the behaviour policy is the same as the optimization policy, or {\it off-policy} if the behaviour policy is different. 

\paragraph{Modality}
We study the use of preference tuning on various modality, such as text, speech, vision, kinesthetic and others if we are not able to classify them.
In the latest advancement of NLP, the idea of RL has been further explored to language and speech tasks, even in multi-modal tasks, such as vision-text. Thus, it is essential to categorize the papers by the extend of the study in terms of the modality, such as text, speech, vision, and vision-text.

\paragraph{Language}
We explore the preference tuning application on different languages. In this case, we categorize the method by English, non-English, and multilingual.

\paragraph{Reward Granularity}

In the preference tuning, the reward can be computed in different granularity levels. The granularity levels can be expanded into two: sample- and token-level. The token-level for each modality may differ, for example, in text tasks, we can use subwords from vocabulary as tokens. And, in vision tasks, patches of image are tokens.

\begin{figure*}[!t]
    \centering
    \includegraphics[width=\linewidth]{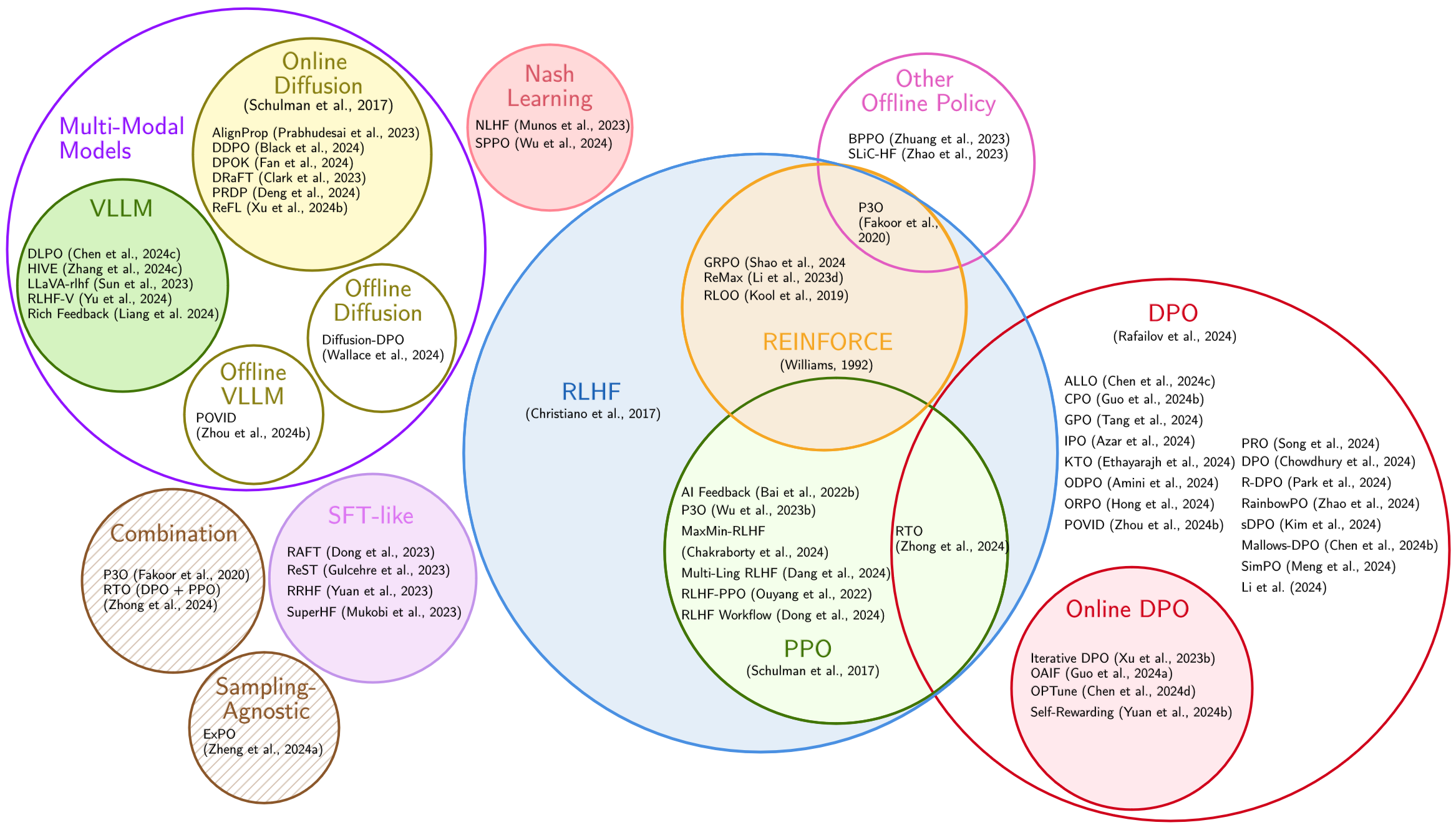} 
    \caption{Preference Tuning methods. The circles with shaded areas represent off-policy methods, while the unshaded circles denote on-policy methods. The overlapping area signifies methods that incorporate both on-policy and off-policy approaches. The policy-agnostic circle indicates methods that are applicable to either on-policy or off-policy scenarios. The combination circle represents methods that integrate both online and off-policy strategies.}
    \label{fig:category}
\end{figure*}

\begin{table*}[!htbp]
\centering
\resizebox{0.93\textwidth}{!}{
    \begin{tabular}{lcccccccccccccccccc}
    \toprule
    \textbf{Method} & \multicolumn{5}{c}{\textbf{Modality}} & \multicolumn{3}{c}{\textbf{Languages}} & \multicolumn{2}{c}{\textbf{Reward Granularity}} \\ \cmidrule{2-11}
    & Text & Speech & Vision & Kinesthetic & Other & EN & Non-EN & Multi. & Sample & Token  \\ \midrule
    \multicolumn{8}{l}{Online Methods}
    \\ \midrule
    RLHF~\citep{christiano2017deep} \\
    $\quad\quad$PPO~\citep{schulman2017proximal} \\
    $\quad\quad\quad\quad$AI Feedback~\citep{bai2022constitutional} & \cellcolor{blue!10}$\checkmark$ & $\times$ & $\times$ & $\times$ & $\times$ & \cellcolor{blue!10}$\checkmark$ & $\times$ & $\times$ & \cellcolor{blue!10}$\checkmark$ & $\times$ \\
    $\quad\quad\quad\quad$P3O~\citep{wu2023pairwise} & \cellcolor{blue!10}$\checkmark$ & $\times$ & $\times$ & $\times$ & $\times$ & \cellcolor{blue!10}$\checkmark$ & $\times$ & $\times$ & \cellcolor{blue!10}$\checkmark$ & $\times$\\
    $\quad\quad\quad\quad$MaxMin-RLHF~\citep{chakraborty2024maxmin} & \cellcolor{blue!10}$\checkmark$ & $\times$ & $\times$ & $\times$ & $\times$ & \cellcolor{blue!10}$\checkmark$ & $\times$ & $\times$ & \cellcolor{blue!10}$\checkmark$ & $\times$  \\
    $\quad\quad\quad\quad$Multi-Ling RLHF~\citep{dang2024rlhf}  & \cellcolor{blue!10}$\checkmark$ & $\times$ & $\times$ & $\times$ & $\times$ & \cellcolor{blue!10}$\checkmark$ & \cellcolor{blue!10}$\checkmark$ & \cellcolor{blue!10}$\checkmark$ &$\times$ &$\times$\\
    $\quad\quad\quad\quad$RLHF-PPO~\citep{ouyang2022training} & \cellcolor{blue!10}$\checkmark$ & $\times$ & $\times$ & $\times$ & $\times$ & \cellcolor{blue!10}$\checkmark$ & $\times$ & $\times$ & \cellcolor{blue!10}$\checkmark$ & $\times$ \\
    $\quad\quad\quad\quad$RLHF Workflow~\citep{dong2024rlhf} & \cellcolor{blue!10}$\checkmark$ & $\times$ & $\times$ & $\times$ & $\times$ & \cellcolor{blue!10}$\checkmark$ & $\times$ & $\times$ & \cellcolor{blue!10}$\checkmark$ & $\times$  \\
    $\quad\quad$REINFORCE~\citep{williams1992simple} \\
    $\quad\quad\quad\quad$GRPO~\citep{shao2024deepseekmath} & \cellcolor{blue!10}$\checkmark$ & $\times$ & $\times$ & $\times$ & $\times$ & \cellcolor{blue!10}$\checkmark$ & \cellcolor{blue!10}$\checkmark$ & \cellcolor{blue!10}$\checkmark$ & \cellcolor{blue!10}$\checkmark$ & $\times$\\ 
    $\quad\quad\quad\quad$ReMax~\citep{li2023remax} & \cellcolor{blue!10}$\checkmark$ & $\times$ & $\times$ & $\times$ & $\times$ & \cellcolor{blue!10}$\checkmark$ & $\times$ & $\times$ & \cellcolor{blue!10}$\checkmark$ & $\times$ \\
    $\quad\quad\quad\quad$RLOO~\citep{ahmadian2024back} & \cellcolor{blue!10}$\checkmark$ & $\times$ & $\times$ & $\times$ & $\times$ & \cellcolor{blue!10}$\checkmark$ & $\times$ & $\times$ & \cellcolor{blue!10}$\checkmark$ & $\times$\\ 
    Online DPO & \\
    $\quad\quad$Iterative DPO~\citep{xu2023some} & \cellcolor{blue!10}$\checkmark$ & $\times$ & $\times$ & $\times$ & $\times$ & \cellcolor{blue!10}$\checkmark$ & $\times$ & $\times$ & \cellcolor{blue!10}$\checkmark$ & $\times$  \\
    $\quad\quad$OAIF~\citep{guo2024direct} & \cellcolor{blue!10}$\checkmark$ & $\times$ & $\times$ & $\times$ & $\times$ & \cellcolor{blue!10}$\checkmark$ & $\times$ & $\times$ & \cellcolor{blue!10}$\checkmark$ & $\times$  \\
    $\quad\quad$OPTune~\citep{chen2024optune} & \cellcolor{blue!10}$\checkmark$ & $\times$ & $\times$ & $\times$ & $\times$ & \cellcolor{blue!10}$\checkmark$ & $\times$ & $\times$ & \cellcolor{blue!10}$\checkmark$ & $\times$  \\
    $\quad\quad$Self-Rewarding \citep{yuan2024self} & \cellcolor{blue!10}$\checkmark$ & $\times$ & $\times$ & $\times$ & $\times$ & \cellcolor{blue!10}$\checkmark$ & $\times$ & $\times$ & \cellcolor{blue!10}$\checkmark$ & $\times$  \\
    Nash-Learning \\
    $\quad\quad$NLHF~\citep{munos2023nash} & \cellcolor{blue!10}$\checkmark$ & $\times$ & $\times$ & $\times$ & $\times$ & \cellcolor{blue!10}$\checkmark$ & $\times$ & $\times$ & \cellcolor{blue!10}$\checkmark$ & $\times$  \\
    $\quad\quad$SPPO~\citep{wu2024self} & \cellcolor{blue!10}$\checkmark$ & $\times$ & $\times$ & $\times$ & $\times$ & \cellcolor{blue!10}$\checkmark$ & $\times$ & $\times$ & \cellcolor{blue!10}$\checkmark$ & $\times$  \\
    SFT-like \\
    $\quad\quad$RAFT~\citep{dong2023raft} & \cellcolor{blue!10}$\checkmark$ & $\times$ & $\times$ & $\times$ & $\times$ & \cellcolor{blue!10}$\checkmark$ & $\times$ & $\times$ & \cellcolor{blue!10}$\checkmark$ & $\times$  \\ 
    $\quad\quad$ReST~\citep{gulcehre2023reinforced}& \cellcolor{blue!10}$\checkmark$ & $\times$ & $\times$ & $\times$ & $\times$ & \cellcolor{blue!10}$\checkmark$ & $\times$ & $\times$ & \cellcolor{blue!10}$\checkmark$ & $\times$  \\ 
    $\quad\quad$RRHF~\citep{yuan2023rrhf}& \cellcolor{blue!10}$\checkmark$ & $\times$ & $\times$ & $\times$ & $\times$ & \cellcolor{blue!10}$\checkmark$ & $\times$ & $\times$ & \cellcolor{blue!10}$\checkmark$ & $\times$  \\ 
    $\quad\quad$SuperHF~\citep{mukobi2023superhf} & \cellcolor{blue!10}$\checkmark$ & $\times$ & $\times$ & $\times$ & $\times$ & \cellcolor{blue!10}$\checkmark$ & $\times$ & $\times$ & \cellcolor{blue!10}$\checkmark$ & $\times$  \\ 
    Multi-Modal Models\\
    $\quad\quad$Diffusion~\citep{schulman2017proximal} \\
    $\quad\quad\quad\quad$AlignProp~\citep{prabhudesai2023aligning} & \cellcolor{blue!10}$\checkmark$ & $\times$ & \cellcolor{blue!10}$\checkmark$ & $\times$ & $\times$ & \cellcolor{blue!10}$\checkmark$ & $\times$ & $\times$ & \cellcolor{blue!10}$\checkmark$ & $\times$ \\
    $\quad\quad\quad\quad$DDPO~\citep{black2024training}& \cellcolor{blue!10}$\checkmark$ & $\times$ & \cellcolor{blue!10}$\checkmark$ & $\times$ & $\times$ & \cellcolor{blue!10}$\checkmark$ & $\times$ & $\times$ & \cellcolor{blue!10}$\checkmark$ & $\times$ \\
    $\quad\quad\quad\quad$DPOK~\citep{fan2024reinforcement} & \cellcolor{blue!10}$\checkmark$ & $\times$ & \cellcolor{blue!10}$\checkmark$ & $\times$ & $\times$ & \cellcolor{blue!10}$\checkmark$ & $\times$ & $\times$ & \cellcolor{blue!10}$\checkmark$ & $\times$ \\
    $\quad\quad\quad\quad$DRaFT~\citep{clark2023directly}& \cellcolor{blue!10}$\checkmark$ & $\times$ & \cellcolor{blue!10}$\checkmark$ & $\times$ & $\times$ & \cellcolor{blue!10}$\checkmark$ & $\times$ & $\times$ & \cellcolor{blue!10}$\checkmark$ & $\times$ \\
    $\quad\quad\quad\quad$PRDP~\citep{deng2024prdp} & \cellcolor{blue!10}$\checkmark$ & $\times$ & \cellcolor{blue!10}$\checkmark$ & $\times$ & $\times$ & \cellcolor{blue!10}$\checkmark$ & $\times$ & $\times$ & \cellcolor{blue!10}$\checkmark$ & $\times$ \\
    $\quad\quad\quad\quad$ReFL~\citep{xu2024imagereward} & \cellcolor{blue!10}$\checkmark$ & $\times$ & \cellcolor{blue!10}$\checkmark$ & $\times$ & $\times$ & \cellcolor{blue!10}$\checkmark$ & $\times$ & $\times$ & \cellcolor{blue!10}$\checkmark$ & $\times$ \\
    $\quad\quad$VLLM~\citep{liu2024visual} \\
    $\quad\quad\quad\quad$DLPO~\citep{chen2024reinforcement} & \cellcolor{blue!10}$\checkmark$ & $\times$ & \cellcolor{blue!10}$\checkmark$ & $\times$ & $\times$ & \cellcolor{blue!10}$\checkmark$ & $\times$ & $\times$ & \cellcolor{blue!10}$\checkmark$ & $\times$ \\
    $\quad\quad\quad\quad$HIVE~\citep{zhang2024hive} & \cellcolor{blue!10}$\checkmark$ & $\times$ & \cellcolor{blue!10}$\checkmark$ & $\times$ & $\times$ & \cellcolor{blue!10}$\checkmark$ & $\times$ & $\times$ & \cellcolor{blue!10}$\checkmark$ & $\times$ \\
    $\quad\quad\quad\quad$LLaVA-rlhf~\citep{sun2023aligning} & \cellcolor{blue!10}$\checkmark$ & $\times$ & \cellcolor{blue!10}$\checkmark$ & $\times$ & $\times$ & \cellcolor{blue!10}$\checkmark$ & $\times$ & $\times$ & \cellcolor{blue!10}$\checkmark$ & $\times$ \\
    $\quad\quad\quad\quad$RLHF-V~\citep{yu2024rlhf}& \cellcolor{blue!10}$\checkmark$ & $\times$ & \cellcolor{blue!10}$\checkmark$ & $\times$ & $\times$ & \cellcolor{blue!10}$\checkmark$ & $\times$ & $\times$ & \cellcolor{blue!10}$\checkmark$ & $\times$ \\
    $\quad\quad\quad\quad$Rich Feedback \citep{liang2024rich}& \cellcolor{blue!10}$\checkmark$ & $\times$ & \cellcolor{blue!10}$\checkmark$ & $\times$ & $\times$ & \cellcolor{blue!10}$\checkmark$ & $\times$ & $\times$ & \cellcolor{blue!10}$\checkmark$ & $\times$ \\
    \midrule
    \multicolumn{8}{l}{Offline Methods} \\ \midrule 
    BPPO~\citep{zhuang2023behavior} & $\times$ & $\times$ & $\times$ & \cellcolor{blue!10}$\checkmark$ & $\times$ & $\times$ & $\times$ & $\times$ & \cellcolor{blue!10}$\checkmark$ & $\times$  \\
    Multi-Modal Models\\
    $\quad\quad$Diffusion-DPO~\citep{wallace2024diffusion} & \cellcolor{blue!10}$\checkmark$ & $\times$ & \cellcolor{blue!10}$\checkmark$ & $\times$ & $\times$ & \cellcolor{blue!10}$\checkmark$ & $\times$ & $\times$ & \cellcolor{blue!10}$\checkmark$ & $\times$ \\
    $\quad\quad$POVID~\citep{zhou2024aligning} & \cellcolor{blue!10}$\checkmark$ & $\times$ & \cellcolor{blue!10}$\checkmark$ & $\times$ & $\times$ & \cellcolor{blue!10}$\checkmark$ & $\times$ & $\times$ & \cellcolor{blue!10}$\checkmark$ & $\times$ \\
    Offline DPO~\citep{rafailov2024direct} & \cellcolor{blue!10}$\checkmark$ & $\times$ & $\times$ & $\times$ & $\times$ & \cellcolor{blue!10}$\checkmark$ & $\times$ & $\times$ & \cellcolor{blue!10}$\checkmark$ & $\times$ \\
    $\quad\quad$ALLO~\citep{chen2024low} & \cellcolor{blue!10}$\checkmark$ & $\times$ & $\times$ & $\times$ & $\times$ & \cellcolor{blue!10}$\checkmark$ & $\times$ & $\times$ & $\times$ &  \cellcolor{blue!10}$\checkmark$ \\
    $\quad\quad$CPO~\citep{guo2024controllable} & \cellcolor{blue!10}$\checkmark$ & $\times$ & $\times$ & $\times$ & $\times$ & \cellcolor{blue!10}$\checkmark$ & \cellcolor{blue!10}$\checkmark$ & \cellcolor{blue!10}$\checkmark$ & \cellcolor{blue!10}$\checkmark$ & $\times$ \\ 
    $\quad\quad$GPO~\citep{tang2024generalized} & \cellcolor{blue!10}$\checkmark$ & $\times$ & $\times$ & $\times$ & $\times$ & \cellcolor{blue!10}$\checkmark$ & $\times$ & $\times$ & \cellcolor{blue!10}$\checkmark$ & $\times$ \\ 
    $\quad\quad$IPO~\citep{azar2024general} & $\times$ & $\times$ & $\times$ & $\times$ & \cellcolor{blue!10}$\checkmark$ & $\times$ & $\times$ & $\times$ & \cellcolor{blue!10}$\checkmark$ & $\times$  \\
    $\quad\quad$KTO~\citep{ethayarajh2024kto} & \cellcolor{blue!10}$\checkmark$ & $\times$ & $\times$ & $\times$ & $\times$ & \cellcolor{blue!10}$\checkmark$ & $\times$ & $\times$ & \cellcolor{blue!10}$\checkmark$ & $\times$  \\
    $\quad\quad$ODPO \citep{amini2024direct}
    & \cellcolor{blue!10}$\checkmark$ & $\times$ & $\times$ & $\times$ & $\times$ & \cellcolor{blue!10}$\checkmark$ & $\times$ & $\times$ & \cellcolor{blue!10}$\checkmark$ & $\times$ \\ 
    $\quad\quad$ORPO~\citep{hong2024orpo} & \cellcolor{blue!10}$\checkmark$ & $\times$ & $\times$ & $\times$ & $\times$ & \cellcolor{blue!10}$\checkmark$ & $\times$ & $\times$ & \cellcolor{blue!10}$\checkmark$ & $\times$ \\
    $\quad\quad$PRO~\citep{song2024preference} & \cellcolor{blue!10}$\checkmark$ & $\times$ & $\times$ & $\times$ & $\times$ & \cellcolor{blue!10}$\checkmark$ & $\times$ & $\times$ & \cellcolor{blue!10}$\checkmark$ & $\times$ \\
    $\quad\quad$R-DPO~\citep{park2024disentangling} & \cellcolor{blue!10}$\checkmark$ & $\times$ & $\times$ & $\times$ & $\times$ & \cellcolor{blue!10}$\checkmark$ & $\times$ & $\times$ & \cellcolor{blue!10}$\checkmark$ & $\times$ \\
    $\quad\quad$rDPO~\citep{chowdhury2024provably} & \cellcolor{blue!10}$\checkmark$ & $\times$ & $\times$ & $\times$ & $\times$ & \cellcolor{blue!10}$\checkmark$ & $\times$ & $\times$ & \cellcolor{blue!10}$\checkmark$ & $\times$ \\
    $\quad\quad$sDPO~\citep{kim2024sdpo} & \cellcolor{blue!10}$\checkmark$ & $\times$ & $\times$ & $\times$ & $\times$ & \cellcolor{blue!10}$\checkmark$ & $\times$ & $\times$ & \cellcolor{blue!10}$\checkmark$ & $\times$  \\
    $\quad\quad$VPO~\citep{chen2024mallows} & \cellcolor{blue!10}$\checkmark$ & $\times$ & $\times$ & $\times$ & $\times$ & \cellcolor{blue!10}$\checkmark$ & $\times$ & $\times$ & \cellcolor{blue!10}$\checkmark$ & $\times$ \\
    $\quad\quad$Mallows-DPO~\citep{chen2024mallows} & \cellcolor{blue!10}$\checkmark$ & $\times$ & $\times$ & $\times$ & $\times$ & \cellcolor{blue!10}$\checkmark$ & $\times$ & $\times$ & \cellcolor{blue!10}$\checkmark$ & $\times$ \\
    $\quad\quad$RainbowPO~\citep{zhao2024rainbowpo} & \cellcolor{blue!10}$\checkmark$ & $\times$ & $\times$ & $\times$ & $\times$ & \cellcolor{blue!10}$\checkmark$ & $\times$ & $\times$ & \cellcolor{blue!10}$\checkmark$ & $\times$ \\
    $\quad\quad$SimPO~\citep{meng2024simpo} & \cellcolor{blue!10}$\checkmark$ & $\times$ & $\times$ & $\times$ & $\times$ & \cellcolor{blue!10}$\checkmark$ & $\times$ & $\times$ & \cellcolor{blue!10}$\checkmark$ & $\times$ \\
    $\quad\quad$\citep{li2024preference} & \cellcolor{blue!10}$\checkmark$ & $\times$ & $\times$ & $\times$ & $\times$ & \cellcolor{blue!10}$\checkmark$ & \cellcolor{blue!10}$\checkmark$ & \cellcolor{blue!10}$\checkmark$ & \cellcolor{blue!10}$\checkmark$ & $\times$ \\
    SLiC-HF~\citep{zhao2023slic} & \cellcolor{blue!10}$\checkmark$ & $\times$ & $\times$ & $\times$ & $\times$ & \cellcolor{blue!10}$\checkmark$ & $\times$ & $\times$ & \cellcolor{blue!10}$\checkmark$ & $\times$  \\
    \midrule
    \multicolumn{8}{l}{Combination} \\ \midrule
    P3O~\citep{fakoor2020p3o} & $\times$ & $\times$ & \cellcolor{blue!10}$\checkmark$ & $\times$ & $\times$ & $\times$ & $\times$ & $\times$ & \cellcolor{blue!10}$\checkmark$ & $\times$  \\
    RTO (DPO + PPO)~\citep{zhong2024dpo} & \cellcolor{blue!10}$\checkmark$ & $\times$ & $\times$ & $\times$ & $\times$ & \cellcolor{blue!10}$\checkmark$ & $\times$ & $\times$ & $\times$ &\cellcolor{blue!10}$\checkmark$\\ \midrule
    \multicolumn{8}{l}{Sampling-Agnostic} \\ \midrule
    ExPO~\citep{zheng2024weak} & \cellcolor{blue!10}$\checkmark$ & $\times$ & $\times$ & $\times$ & $\times$ & \cellcolor{blue!10}$\checkmark$ & $\times$ & $\times$ & \cellcolor{blue!10}$\checkmark$ & $\times$ \\
    \bottomrule
    \end{tabular}
}
\caption{Preference Tuning methods. The categorization based on the methods under study and it does not limit the extension of the method to other domains or modalities.}
\label{tab:category}
\end{table*}

\section{Preference Tuning}
\begin{figure*}[!t]
    \centering
    \includegraphics[width=0.87\linewidth]{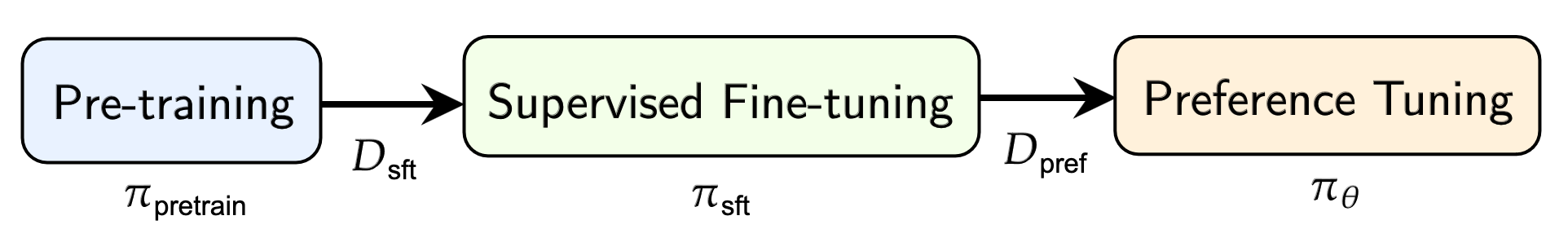} 
    \caption{Training stages.}
    \label{fig:training-stages}
\end{figure*}

In this section, we cover the general framework to train preference-tuned generative models. As shown in Table~\ref{fig:training-stages}, the preference tuning training framework typically begins with the supervised fine-tuning (SFT) stage, during which the generative model is trained to excel at next-token prediction or use an instruction-tuned model as the base initialized model. The SFT focuses on improving the model capability to generate tokens as it guides the model on how an generative model should response to a prompt input. Once the model is able to properly generate fluent text sequences, the model is further aligned by further policy optimization via RL. The alignment is useful to guide the model to answer with a appropriate manner based on the preference objective. This step is a necessary training stage to make sure the model generation aligned to human preference, thus, the model will act more human-like. Notably, the human alignment stage can also be jointly trained alongside SFT.
\begin{figure*}[!t]
    \centering
    \includegraphics[width=\linewidth]{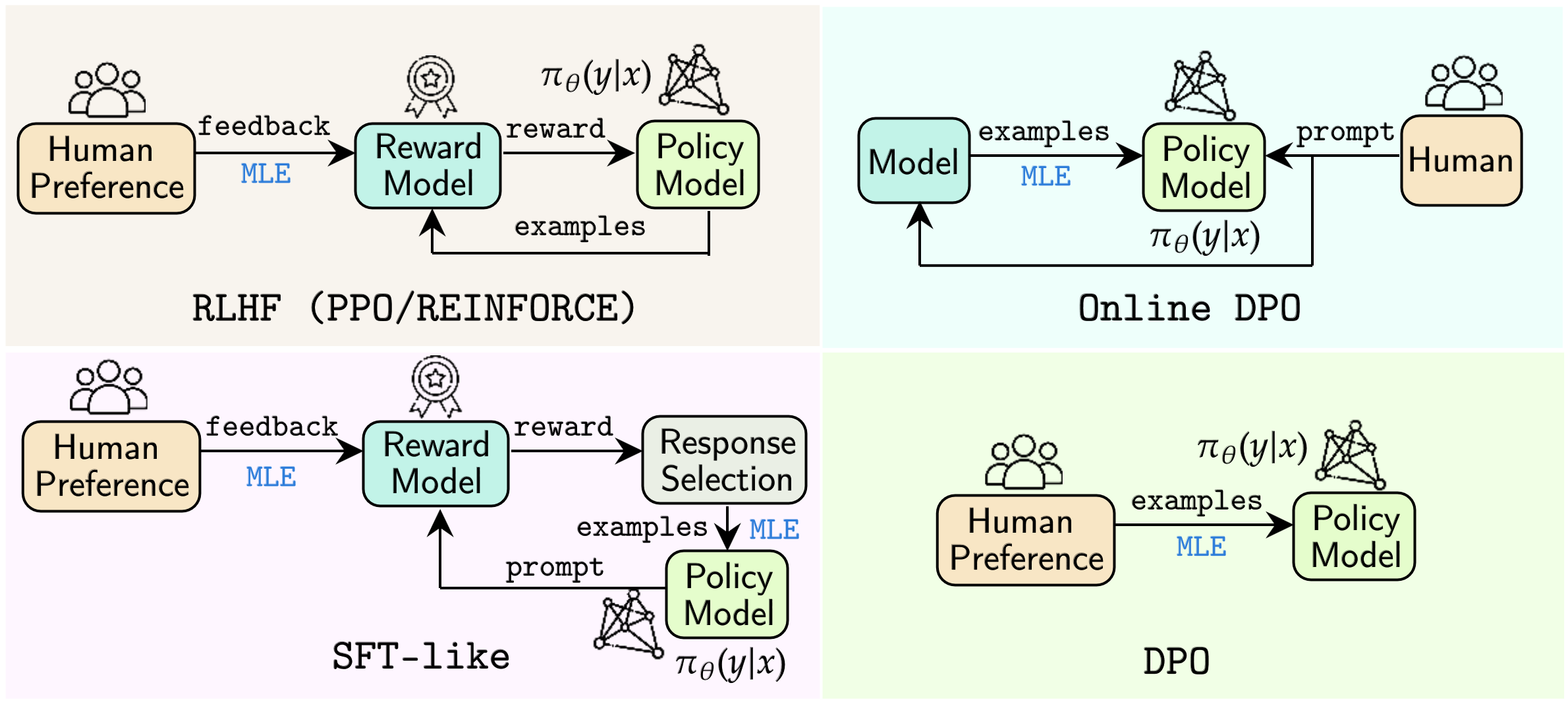} 
    \caption{Preference Tuning methods for online algorithms, such as RLHF, Online DPO, and SFT-like, and offline methods, such as DPO.}
    \label{fig:diagram}
\end{figure*}
\subsection{Training Phases}
The training phases for preference tuning are described as follows.
\subsubsection{Supervised Fine-Tuning (SFT)}
On the preference tuning, a generative model with trainable weights $\theta$ normally starts by SFT via maximum likelihood (MLE) using teacher forcing and cross-entropy loss. The training is done using the supervised fine-tuning dataset $\mathcal{D}_\text{sft}$. The objective is to maximize the log probability of a set of human demonstrations. The generative model is trained to generate the label by predicting the next token $y_{t+1}$ given the input $x$, current and previous label tokens $y_{t:<t}$. During the SFT, we utilize an attention mask applying to the entire context $x$ and $y_{t:<t}$, and avoid applying attention to future tokens. The trained model denoted $\pi_
\theta^{\text{sft}}$ and it is often to be used to initialize reward model and policy model $\pi_\theta$.

\subsubsection{Reward Modeling}
The reward model $r_\phi(x,y)$ can be trained either separately (offline) or jointly trained with the policy model $\pi_\theta$ (online). Table~\ref{tab:reward-models} shows the list of reward models.

\begin{table*}[!t]
\centering
\resizebox{0.95\textwidth}{!}{
    \begin{tabular}{lccc}
    \toprule
    \textbf{Reward Model} & \textbf{Sizes} & \textbf{Model Base} & \textbf{Datasets} \\ \midrule
    Single Objective \\ \midrule
    BTRM Qwen2 & 7B\href{https://huggingface.co/CIR-AMS/BTRM_Qwen2_7b_0613}{$^{\triangle}$} & Qwen2 & UNK \\
    
Eurus-RM~\citep{yuan2024advancing} & 7B\href{https://huggingface.co/openbmb/Eurus-RM-7b}{$^{\triangle}$} & Mistral & UltraInteract, UltraFeedback, UltraSafety \\
    FsfairX-LLama3-v0.1~\citep{dong2023raft} & 8B\href{https://huggingface.co/sfairXC/FsfairX-LLaMA3-RM-v0.1}{$^{\triangle}$} & Llama3 & UNK \\
    GRM-llama3-8B-sftreg~\citep{yang2024regularizing} & 8B\href{https://huggingface.co/Ray2333/GRM-llama3-8B-sftreg}{$^{\triangle}$} & Llama3 & Preference 700K \\
    GRM-llama3-8B-distill~\citep{yang2024regularizing} & 8B\href{https://huggingface.co/Ray2333/GRM-llama3-8B-distill}{$^{\triangle}$} & Llama3 & Preference 700K \\
    InternLM2~\citep{cai2024internlm2} & 1.8B\href{https://huggingface.co/internlm/internlm2-1_8b-reward}{$^{\triangle}$}, 7B\href{https://huggingface.co/internlm/internlm2-20b-reward}{$^{\triangle}$}, 20B\href{https://huggingface.co/internlm/internlm2-20b-reward}{$^{\triangle}$}  & UNK & UNK \\
    SteerLM-Llama3~\citep{wang2024helpsteer2} & 70B\href{https://huggingface.co/nvidia/Llama3-70B-SteerLM-RM}{$^{\triangle}$} & Llama3 & HelpSteer2\\
    Nemotron-4-340B-Reward~\citep{adler2024nemotron} & 340B\href{https://huggingface.co/nvidia/Nemotron-4-340B-Reward}{$^{\triangle}$} & Nemotron4 & HelpSteer2 \\
    Pair-preference-model-LLamA3-8B~\citep{dong2024rlhf} & 8B\href{https://huggingface.co/RLHFlow/pair-preference-model-LLaMA3-8B}{$^{\triangle}$} & LLama3 & RLHFlow Pair Preference \\
    Starling-RM-34B & 34B\href{https://huggingface.co/Nexusflow/Starling-RM-34B}{$^{\triangle}$} & Yi-34B-Chat & Nectar \\
    UltraRM~\citep{cui2023ultrafeedback} & 13B\href{https://huggingface.co/openbmb/UltraRM-13b}{$^{\triangle}$} & Llama2 & UltraFeedback \\
    \midrule
    Multi-Objective \\ \midrule
    ArmoRM-Llama3-8B-v0.1~\citep{wang2024interpretable} & 8B\href{https://huggingface.co/RLHFlow/ArmoRM-Llama3-8B-v0.1}{$^{\triangle}$} & Llama3 & HelpSteer, UltraFeedback, BeaverTails-30k \\
    & & & CodeUltraFeedback, Prometheus, Argilla-Capybara \\
    & & & Argilla-OpenOrca, Argilla-Math-Preference
    \\ \midrule
    Multi-Model \\
    \midrule
     MetaMetrics-RM~\citep{winata2024metametrics} &  Multiple & Multiple & Skywork Preference Data and AllenAI Preference Data
    \\ 
     \\ \bottomrule
    \end{tabular}
}
\caption{Reward Models.}
\label{tab:reward-models}
\end{table*}

\paragraph{Single Objective Reward Model} Bradley-Terry Reward Model~\citep{bradley1952rank} is a pairwise comparison between two samples. It estimates the probability that the pairwise comparison $i \succ j$, which indicates a strong preference of $i$ over $j$, is true as:
\begin{align}
\label{BT}
    P(i \succ j) = \frac{\exp{s_i}}{\exp{s_i} + \exp{s_j}},
\end{align}
where $s_i$ and $s_j$ are latent variables representing sample $i$ and sample $j$, respectively. Thus, given the preference dataset $\mathcal{D}_\text{pref} = \{x^i, y_w^i, y_l^i\}_{i=1}^{N}$, we could obtain an estimation of the reward model $r_\phi(x,y)$ by minimizing the negative log-likelihood loss:
\begin{align}
   \mathcal{L}(\phi) &= -\mathbb{E}_{(x, y_w, y_l)\sim\mathcal{D}_\text{pref}}\log P(y_w\succ y_l\mid x)\\ 
   &= -\mathbb{E}_{(x, y_w, y_l)\sim\mathcal{D}_\text{pref}}\log \sigma(r_{\phi}(x,y_w)-r_{\phi}(x,y_l)),
\end{align}
which $\sigma$ denotes the logistic function, i.e., $\sigma(x):=(1+e^{-x})^{-1}$.

\paragraph{Multi-Objective Reward Model}
Absolute-Rating Multi-Objective Reward Model (ArmoRM)~\citep{wang2024interpretable} is a two-stage approach that first trains a multi-objective RM and then learns a gating layer that scalarizes reward objectives in a mixture-of-experts way. Each example consists of an input $x$ and output $y$ with $k$-dimensional rating vector, where each dimension corresponds to a reward objective. A concatenation of input and output $x\oplus y$ is passed through the model $f_\theta$ with a linear regression layer $w$, which outputs a $k$-dimensional rating prediction. The model is trained with regression loss:
\begin{align}
\min_{\theta, w} \mathbb{E}_{x,y,r\in \mathcal{D}}\|w^\top f_\theta(x\oplus y) - r\|_2^2.
\end{align}
Then, it learns a mixture-of-experts gating function, $g_\phi$, which is implemented as a shallow MLP. This MLP takes the representation of the input $x$ and outputs a $k$-dimensional vector, which is then processed by a softmax function. During the training of the gating layer, the backbone and the regression layer are kept frozen. Only the gating layer is trained using the Bradley-Terry loss, augmented with an additional scaling variable.

\paragraph{Multi-Model Reward Model} MetaMetrics~\citep{winata2024metametrics} is a method to combine multiple existing reward models into a more powerful reward model by calibrating them using the preference data. The method is a systematic way to identify reward models that can be used complementary without blindly use the models. There are two methods introduced to calibrate the models using Bayesian optimization and boosting method. Thus, the approach is highly efficient and they are aspect-agnostic, thus allowing flexibility to use them in any preference data. 

\subsubsection{Preference Alignment using Reinforcement Learning}
While SFT has led to markedly improved performance, there is still a misalignment between SFT objective and the ultimate target of generating high-quality outputs as determined by humans. \cite{stiennon2020learning,ouyang2022training} propose reinforcement learning from human feedback (RLHF) to further align language models with human intent. RLHF pipeline starts with the stage of modeling the rewards from human preferences, known as reward modeling stage, by maximizing the likelihood of preferences under the ground truth assumption. After obtaining the RM, RLHF further trains the Language Model policy via Reinforcement Learning to maximize the score given by the RM. Proximal Policy Optimization (PPO) was commonly chosen as the RL algorithm to update the policy because of its great sample efficiency.

\subsubsection{Joint Training}
Recent works also proposed that two stages of SFT and RLHF can be simplied as one stage with a weighted combination of the two loss functions and even lead to better performance. The key takeaway is to treat the preferred answer in the Human Alignment/RLHF stage as the SFT target, e.g., SLiC-HF~\citep{zhao2023slic}.

\begin{table*}[!t]
\centering
\resizebox{0.98\textwidth}{!}{
    \begin{tabular}{lccccccccc}
    \toprule
    \textbf{Dataset} & \textbf{\# Samples}  & \multicolumn{2}{c}{\textbf{Usecase}} & \multicolumn{2}{c}{\textbf{Data Source}} & \multicolumn{1}{c}{\textbf{Annotation}} \\ \cmidrule{3-7}
    & \textbf{or (\# Tokens) or [Byte Size]} & \textbf{SFT} & \textbf{Alignment} & \textbf{Human} & \textbf{LLM} & \textbf{Human}\\ \midrule
    Alpaca~\citep{taori2023stanford} & 52k & \cellcolor{blue!10}$\checkmark$ & $\times$ & \cellcolor{blue!10}$\checkmark$ & \cellcolor{blue!10}$\checkmark$ & $\times$ \\
    Alpaca-CoT\href{https://github.com/PhoebusSi/Alpaca-CoT}{$^{\triangle}$} & 127.5M$^\dagger$ & \cellcolor{blue!10}$\checkmark$ & $\times$ & \cellcolor{blue!10}$\checkmark$ & \cellcolor{blue!10}$\checkmark$ & \cellcolor{blue!10}$\checkmark$ \\
    Aya Dataset~\citep{singh2024aya} & 202k & \cellcolor{blue!10}$\checkmark$ & $\times$ & \cellcolor{blue!10}$\checkmark$ & $\times$ & \cellcolor{blue!10}$\checkmark$ \\
    ChatAlpaca\href{https://github.com/cascip/ChatAlpaca}{$^{\triangle}$} & 20k$^\dagger$ & \cellcolor{blue!10}$\checkmark$ & $\times$ & \cellcolor{blue!10}$\checkmark$ & \cellcolor{blue!10}$\checkmark$ & $\times$\\
    BeaverTails~\citep{ji2024beavertails} & 30k, 330k & \cellcolor{blue!10}$\checkmark$ & \cellcolor{blue!10}$\checkmark$ & \cellcolor{blue!10}$\checkmark$ & \cellcolor{blue!10}$\checkmark$ & \cellcolor{blue!10}$\checkmark$  \\
    Code-Alpaca\href{https://github.com/sahil280114/codealpaca}{$^{\triangle}$} & 20k & \cellcolor{blue!10}$\checkmark$ & $\times$ & \cellcolor{blue!10}$\checkmark$ & \cellcolor{blue!10}$\checkmark$ & $\times$ \\
    CodeUltraFeedback\href{https://huggingface.co/datasets/coseal/CodeUltraFeedback}{$^{\triangle}$} & 10k & \cellcolor{blue!10}$\checkmark$ & \cellcolor{blue!10}$\checkmark$ & \cellcolor{blue!10}$\checkmark$ & \cellcolor{blue!10}$\checkmark$ & \cellcolor{blue!10}$\checkmark$ \\
    Dolly~\citep{conover2023free} & 15k & \cellcolor{blue!10}$\checkmark$ & $\times$ & \cellcolor{blue!10}$\checkmark$ & $\times$ & \cellcolor{blue!10}$\checkmark$\\
    FLAN collection~\citep{longpre2023flan} & UNK$^\ddagger$ & \cellcolor{blue!10}$\checkmark$ & $\times$ & \cellcolor{blue!10}$\checkmark$ & $\times$ & \cellcolor{blue!10}$\checkmark$ \\
    HC3~\citep{guo2023close} & 24.3k & \cellcolor{blue!10}$\checkmark$ & \cellcolor{blue!10}$\checkmark$ & \cellcolor{blue!10}$\checkmark$ & \cellcolor{blue!10}$\checkmark$ & \cellcolor{blue!10}$\checkmark$\\
    HelpSteer2~\citep{wang2024helpsteer2} & 21k & \cellcolor{blue!10}$\checkmark$ & \cellcolor{blue!10}$\checkmark$ & \cellcolor{blue!10}$\checkmark$ & \cellcolor{blue!10}$\checkmark$ & \cellcolor{blue!10}$\checkmark$\\
    HH-RLHF~\citep{bai2022training} & 170k & $\times$ & \cellcolor{blue!10}$\checkmark$ & \cellcolor{blue!10}$\checkmark$ & $\times$ & \cellcolor{blue!10}$\checkmark$  \\
    InstructionWild v2~\citep{ni2023instruction} & 110k & \cellcolor{blue!10}$\checkmark$ & $\times$ & \cellcolor{blue!10}$\checkmark$ & $\times$ & \cellcolor{blue!10}$\checkmark$ \\
    LIMA~\citep{zhou2024lima} & 1.3k & \cellcolor{blue!10}$\checkmark$ & $\times$ & \cellcolor{blue!10}$\checkmark$ & $\times$ & \cellcolor{blue!10}$\checkmark$ \\
    Magpie (Air)~\citep{xu2024magpie} & 300k, 3M & \cellcolor{blue!10}$\checkmark$ & \cellcolor{blue!10}$\checkmark$ & \cellcolor{blue!10}$\checkmark$ & \cellcolor{blue!10}$\checkmark$ & \cellcolor{blue!10}$\checkmark$\\ 
    Magpie (Pro)~\citep{xu2024magpie} & 300k, 1M & \cellcolor{blue!10}$\checkmark$ & \cellcolor{blue!10}$\checkmark$ & \cellcolor{blue!10}$\checkmark$ & \cellcolor{blue!10}$\checkmark$ & \cellcolor{blue!10}$\checkmark$\\ 
    M2Lingual~\citep{maheshwary2024m2lingual} & 174k & \cellcolor{blue!10}$\checkmark$ & $\times$ &  \cellcolor{blue!10}$\checkmark$ &  \cellcolor{blue!10}$\checkmark$ & $\times$ \\
    Natural Questions~\citep{kwiatkowski2019natural} & 323k & \cellcolor{blue!10}$\checkmark$ & $\times$ & \cellcolor{blue!10}$\checkmark$ & $\times$ & \cellcolor{blue!10}$\checkmark$ \\
    Oasst1~\citep{kopf2024openassistant} & 88.8k & \cellcolor{blue!10}$\checkmark$ & \cellcolor{blue!10}$\checkmark$ & \cellcolor{blue!10}$\checkmark$ & $\times$ & \cellcolor{blue!10}$\checkmark$ \\
    Okapi~\citep{lai2023okapi} & 4.3M$^*$ & \cellcolor{blue!10}$\checkmark$ & \cellcolor{blue!10}$\checkmark$ & \cellcolor{blue!10}$\checkmark$ & \cellcolor{blue!10}$\checkmark$ & $\times$\\
    P3~\citep{sanh2021multitask} & 122M & \cellcolor{blue!10}$\checkmark$ & $\times$ & \cellcolor{blue!10}$\checkmark$ & $\times$ & \cellcolor{blue!10}$\checkmark$ \\
    Preference 700K\href{https://huggingface.co/datasets/hendrydong/preference_700K}{$^{\triangle}$} & 700K & $\times$ & \cellcolor{blue!10}$\checkmark$ & UNK & UNK & UNK  \\
    Prometheus2~\citep{kim2024prometheus} & 200k & \cellcolor{blue!10}$\checkmark$ & \cellcolor{blue!10}$\checkmark$ & \cellcolor{blue!10}$\checkmark$ & \cellcolor{blue!10}$\checkmark$ & $\times$\\
    Prosocial-Dialog~\citep{kim2022prosocialdialog} & 165.4k & \cellcolor{blue!10}$\checkmark$ & \cellcolor{blue!10}$\checkmark$ & \cellcolor{blue!10}$\checkmark$ & $
    \times$ & \cellcolor{blue!10}$\checkmark$ \\
    RLHFlow Pair Preference\href{https://huggingface.co/datasets/RLHFlow/pair_preference_model_dataset}{$^{\triangle}$} & 700k & $\times$ & \cellcolor{blue!10}$\checkmark$ & \cellcolor{blue!10}$\checkmark$ & \cellcolor{blue!10}$\checkmark$ & \cellcolor{blue!10}$\checkmark$ \\
    Self-instruct~\citep{wang2023self} & 197k & \cellcolor{blue!10}$\checkmark$ & $\times$ & \cellcolor{blue!10}$\checkmark$ & $\times$ & \cellcolor{blue!10}$\checkmark$ \\
    ShareGPT & Multiple Versions & \cellcolor{blue!10}$\checkmark$ & \cellcolor{blue!10}$\checkmark$ & \cellcolor{blue!10}$\checkmark$ & \cellcolor{blue!10}$\checkmark$ & \cellcolor{blue!10}$\checkmark$ \\
    StackExchange\href{https://huggingface.co/datasets/HuggingFaceH4/stack-exchange-preferences}{$^{\triangle}$} & 10.8M & \cellcolor{blue!10}$\checkmark$ & \cellcolor{blue!10}$\checkmark$ & \cellcolor{blue!10}$\checkmark$ & $\times$ & \cellcolor{blue!10}$\checkmark$ \\ 
    Super-Natural Instructions~\citep{wang2022super} & 5M & \cellcolor{blue!10}$\checkmark$ & \cellcolor{blue!10}$\checkmark$ & \cellcolor{blue!10}$\checkmark$ & \cellcolor{blue!10}$\checkmark$ & \cellcolor{blue!10}$\checkmark$ \\
    UltraChat~\citep{ding2023enhancing} & 1.5M & \cellcolor{blue!10}$\checkmark$ & $\times$ & $\times$ & \cellcolor{blue!10}$\checkmark$ & \cellcolor{blue!10}$\checkmark$ \\
    UltraFeedback~\citep{cui2023ultrafeedback} & 64k & \cellcolor{blue!10}$\checkmark$ & \cellcolor{blue!10}$\checkmark$ & \cellcolor{blue!10}$\checkmark$ & \cellcolor{blue!10}$\checkmark$ & \cellcolor{blue!10}$\checkmark$\\
    WildChat~\citep{zhao2024wildchat} & 652k & \cellcolor{blue!10}$\checkmark$ & \cellcolor{blue!10}$\checkmark$ & \cellcolor{blue!10}$\checkmark$ & \cellcolor{blue!10}$\checkmark$ & \cellcolor{blue!10}$\checkmark$ \\
    WizardLM~\citep{xu2023wizardlm} & 250k & \cellcolor{blue!10}$\checkmark$ & \cellcolor{blue!10}$\checkmark$ & \cellcolor{blue!10}$\checkmark$ & \cellcolor{blue!10}$\checkmark$ & \cellcolor{blue!10}$\checkmark$ \\ 
    xP3~\citep{muennighoff2023crosslingual} & 78.8M & \cellcolor{blue!10}$\checkmark$ & $\times$ & \cellcolor{blue!10}$\checkmark$ & $\times$ & \cellcolor{blue!10}$\checkmark$ \\\bottomrule
    \end{tabular}
}
\caption{SFT and alignment text datasets. $^\dagger$The dataset is updated over the time and the number placed on the table is from the latest dataset released by the authors. $^\ddagger$The exact size is unknown and some the datasets are no longer accessible. $^*$The estimated number of translated and English instructions.}
\label{tab:text-datasets}
\end{table*}

\begin{table*}[!t]
\centering
\resizebox{0.98\textwidth}{!}{
    \begin{tabular}{lccccccccc}
    \toprule
    \textbf{Dataset} & \textbf{\# Samples}  & \multicolumn{2}{c}{\textbf{Usecase}} & \multicolumn{2}{c}{\textbf{Data Source}} & \multicolumn{1}{c}{\textbf{Annotation}} \\ \cmidrule{3-7}
    & \textbf{or (\# Tokens) or [Byte Size]} & \textbf{SFT} & \textbf{Alignment} & \textbf{Human} & \textbf{LLM} & \textbf{Human}\\ \midrule
    ImageRewardDB \citep{xu2024imagereward} & 137k+ & \cellcolor{blue!10}$\checkmark$ & \cellcolor{blue!10}$\checkmark$ & \cellcolor{blue!10}$\checkmark$ & $\times$ & \cellcolor{blue!10}$\checkmark$
    \\
    Pick-a-pic \citep{kirstain2023pick} & 500k+ & $\times$ & \cellcolor{blue!10}$\checkmark$ & \cellcolor{blue!10}$\checkmark$ & \cellcolor{blue!10}$\checkmark$ & \cellcolor{blue!10}$\checkmark$
    \\
    RichHF-18K \citep{liang2024rich} & 18k & $\times$ & \cellcolor{blue!10}$\checkmark$ & \cellcolor{blue!10}$\checkmark$ & $\times$ & \cellcolor{blue!10}$\checkmark$
    \\
    \bottomrule
    \end{tabular}
}
\caption{SFT and alignment vision datasets. $^\dagger$The dataset is updated over the time and the number placed on the table is from the latest dataset released by the authors. $^\ddagger$The exact size is unknown and some the datasets are no longer accessible. $^*$The estimated number of translated and English instructions.}
\label{tab:vision-datasets}
\end{table*}

\subsection{Datasets}
The dataset sources for SFT and preference tuning can be collected from various sources, such as human and LLMs feedback. Table~\ref{tab:text-datasets} shows the list of SFT and alignment text data labeled by the data source either they are collected by human or synthetically generated by LLM.

\subsubsection{SFT Datasets}
The SFT data is useful for training LM on high-quality input-output demonstration pairs. This is usually conducted for the foundation model as initialization. The SFT data can be in the form of prompts with various format.

\paragraph{LLM-Generated Datasets}
\cite{taori2023stanford} propose Alpaca, a dataset with demonstrations generated using OpenAI's GPT-3 text-davinci-003 model. The instruction data can be used to conduct instruction tuning for LLMs and allow them to follow instruction better. A version of Alpaca dataset with Chain-of-Thought (CoT)~\citep{wei2022chain} and it is introduced to further improve the LLM's reasoning ability. Multi-turn datasets generated using LLMs are also created, such as ChatAlpaca, UltraChat~\citep{ding2023enhancing}, and WildChat~\citep{zhao2024wildchat}. 

\paragraph{Human-Generated and Human-Annotated Datasets}
Using human-generated and human-annotated data are essential in training high-quality models. \citet{zhou2024lima} has shown quality is more important than quantity, as shown as using LIMA datasets that models trained only consist of 1,000 carefully human curated prompts and responses, without any reinforcement learning or human preference modeling can outperform models with much larger instruction-tuned datasets.

\paragraph{Dataset Collection}
FLAN collection~\citep{longpre2023flan} is introduced to train a collection of tasks on top of T5 and PaLM models~\citep{raffel2020exploring}. For training multilingual LMs, Cendol Collection~\citep{cahyawijaya2024cendol}, ROOTS~\citep{laurenccon2022bigscience}, and xP3~\citep{muennighoff2023crosslingual} are used in SFT. Other potential datasets are crowd-sourcing datasets, although they are designed for SFT, but they can be useful resources for SFT, such as NusaCrowd~\citep{cahyawijaya2023nusacrowd} and SEACrowd~\citep{lovenia2024seacrowd}.

\subsubsection{Human Preference Alignment Datasets}

The human alignment data can be in the form of pair-wise or ranking format. We can have a set of preferred and dispreferred data $\mathcal{D}_\text{pref}$ for each input sample. For pairwise dataset, we collect pairs of preferred response $y_w$ and dispreferred response $y_l$. In case of multiple responses, we can gather responses $y_0$, $y_1$, $y_2$, $\ldots$ and ask humans to pick the best $y_i$ from each. These datasets have been used to train reward models.

\paragraph{Conversational Datasets} 
Several existing conversational datasets are instrumental in evaluating the quality of dialogue system or chatbot responses. Notable examples include HelpSteer2~\citep{wang2024helpsteer2} and UltraFeedback~\citep{cui2023ultrafeedback}. HelpSteer2 provides alignment scores across five different aspects—helpfulness, correctness, coherence, complexity, and verbosity—collected from human evaluators. UltraFeedback offers alignment scores for four aspects: instruction-following, truthfulness, honesty, and helpfulness. Additionally, HH-RLHF~\citep{bai2022training} introduces datasets labeled with scores for helpfulness and harmlessness.

\paragraph{Code Datasets} 
CodeUltraFeedback comprises 10,000 coding instructions, each annotated with four responses generated by a diverse pool of 14 LLMs~\citep{weyssow2024codeultrafeedback}. These responses are ranked based on five distinct coding preferences: instruction-following, complexity, style, readability, and another instance of instruction-following. The rankings are determined using GPT-3.5 as a judge, providing both numerical scores and detailed textual feedback.

\begin{table*}[!t]
\centering
\resizebox{0.98\textwidth}{!}{
    \begin{tabular}{llccccccc}
    \toprule
    \textbf{Model}  & \textbf{Sizes} & \textbf{SFT/Pref. Tuning Langs.$^\dagger$} & \multicolumn{1}{c}{\textbf{Model Base}} & \multicolumn{1}{c}{\textbf{SFT}} & \multicolumn{1}{c}{\textbf{Pref. Tuning}} \\ 
    \midrule
    \multicolumn{5}{l}{Open-source LM} \\ \midrule
    Aya-23~\citep{aryabumi2024aya} & 8B, 35B & Multi. (23) & Dec-Only; Command R & \cellcolor{blue!10}$\checkmark$ & $\times$ \\
    Aya-101~\citep{ustun2024aya} & 13B & Multi. (101) & Enc-Dec; mT5 & \cellcolor{blue!10}$\checkmark$ & $\times$ \\
    Bactrian-X~\citep{li2023bactrian} & 7B & Multi. (52) & Dec-Only; Llama1 & \cellcolor{blue!10}$\checkmark$ & $\times$\\ 
    BART~\citep{lewis2020bart} & 139M, 406M & English & Enc-Dec & $\times$ & $\times$ \\
    BLOOM~\citep{le2023bloom} & 560M, 1.1B, 1.7B, 3B, 7.1B, 176B & Multi. (46) + Code (13) & Dec-Only & \cellcolor{blue!10}$\checkmark$ & $\times$ \\
    BLOOMZ~\citep{muennighoff2023crosslingual} & 560M, 1.1B, 1.7B, 3B, 7.1B, 176B & Multi. (108) + Code (13) & Dec-Only; BLOOM & \cellcolor{blue!10}$\checkmark$ & $\times$ \\
    Cendol~\citep{cahyawijaya2024cendol} & 7B, 13B & Multi. (10) & Dec-Only; Llama2 & \cellcolor{blue!10}$\checkmark$ & $\times$ \\
    & 300M, 580M, 1.2B, 3.7B, 13B & Multi. (10) & Enc-Dec; mT5 & \cellcolor{blue!10}$\checkmark$ & $\times$ \\
    FLAN-T5~\citep{longpre2023flan} & 80M, 250M, 780M, 3B, 11B & English & Enc-Dec; T5 & \cellcolor{blue!10}$\checkmark$ & $\times$\\
    Llama1~\citep{touvron2023llama} & 6.7B, 13B, 32.5B, 65.2B & English & Dec-Only & $\times$ & $\times$ \\
    M2M-100~\citep{fan2021beyond} & 418M, 1.2B, 12B & Multi. (100) & Enc-Dec & \cellcolor{blue!10}$\checkmark$ & $\times$\\
    mBART~\citep{liu2020multilingual} & 406M & Multi. (25), Multi. (50) & Enc-Dec & \cellcolor{blue!10}$\checkmark$ & $\times$ \\
    Megatron-LM~\citep{shoeybi2019megatron} & 1.2B, 2.5B, 4.2B, 8.3B & English & Dec-Only; GPT-2 & $\times$ & $\times$\\
    MPT (Instruct/Chat)\href{https://github.com/mosaicml/llm-foundry}{$^{\triangle}$} & 7B, 30B & English & Dec-Only & \cellcolor{blue!10}$\checkmark$ & \cellcolor{blue!10}$\checkmark$ \\
    mT0~\citep{muennighoff2023crosslingual} & 560M, 1B7, 3B, 7B1 & Multi. (108) + Code (13) & Enc-Dec; mT5; & \cellcolor{blue!10}$\checkmark$ & $\times$\\
    OLMo~\citep{groeneveld2024olmo} & 1B, 7B & English + Code & Dec-Only & $\times$ & $\times$ \\ 
    OPT~\citep{zhang2022opt} & 125M, 350M, 1.3B, 2.7B, 6.7B, & English & Dec-Only; Megatron-LM & $\times$ & $\times$ \\
    & 13B, 30B, 66B, 175B \\
    Phi1~\citep{gunasekar2023textbooks} & 1.3B & English & Dec-Only & $\times$ & $\times$ \\
    Phi1.5~\citep{li2023textbooks} & 1.3B & English & Dec-Only & $\times$ & $\times$ \\
    Pythia~\citep{biderman2023pythia} & 70M, 160M, 410M, 1B, 1.4B, & English & Decoder-Only; GPT-NeoX & $\times$ & $\times$\\
    & 2.8B, 6.9B, 12B \\
    SantaCoder~\citep{allal2023santacoder} & 1.1B & Code (3) & Dec-Only & \cellcolor{blue!10}$\checkmark$ & $\times$ \\
    StarCoder~\citep{li2023starcoder} & 15.5B & Code (80+) & Dec-Only & \cellcolor{blue!10}$\checkmark$ & $\times$ \\
    T0~\citep{sanh2021multitask} & 3B, 11B & English & Enc-Dec; T5 & \cellcolor{blue!10}$\checkmark$ & $\times$ \\
    T5~\citep{raffel2020exploring} & 80M, 250M, 780M, 3B, 11B & English & Enc-Dec & \cellcolor{blue!10}$\checkmark$ & $\times$ \\
    T5v1.1~\citep{raffel2020exploring,shazeer2020glu} & 80M, 250M, 780M, 3B, 11B & English & Enc-Dec & $\times$ & $\times$ \\
    WizardCoder~\citep{luo2023wizardcoder} & 7B, 13B, 15B, 33B & Code & Dec-Only & \cellcolor{blue!10}$\checkmark$ & $\times$ \\
\midrule
    \multicolumn{5}{l}{Open-weight LM} \\ \midrule
    
    Alpaca\href{https://crfm.stanford.edu/2023/03/13/alpaca.html}{$^{\triangle}$} & 7B & English & Dec-Only; Llama1 & \cellcolor{blue!10}$\checkmark$ & $\times$ \\ 
    C4AI Command-R (incl. Plus)\href{https://huggingface.co/CohereForAI/c4ai-command-r-v01}{$^{\triangle}$} & 35B, 104B & Multi. (13) & Dec-Only & \cellcolor{blue!10}$\checkmark$ & \cellcolor{blue!10}$\checkmark$ \\
    DBRX\href{https://huggingface.co/databricks/dbrx-instruct}{$^{\triangle}$} & 132B & Multi. (UNK) + Code & MoE & \cellcolor{blue!10}$\checkmark$ & \cellcolor{blue!10}$\checkmark$ \\ 
    DeepSeek-V2\href{https://github.com/deepseek-ai/DeepSeek-V2}{$^{\triangle}$} & 16B, 236B & Multi. (UNK) + Code & MoE & \cellcolor{blue!10}$\checkmark$ & \cellcolor{blue!10}$\checkmark$ \\ 
    Falcon~\citep{almazrouei2023falcon} & 7B, 40B, 180B & Multi. (2) + Code & Dec-Only & \cellcolor{blue!10}$\checkmark$ & $\times$ \\
    Falcon2\href{https://huggingface.co/blog/falcon2-11b}{$^{\triangle}$} & 11B & Multi. (11) + Code & Dec-Only & \cellcolor{blue!10}$\checkmark$ & $\times$\\
    Gemma~\citep{team2024gemma} & 2B, 7B & Multi. (UNK) + Code & Dec-Only & \cellcolor{blue!10}$\checkmark$ & \cellcolor{blue!10}$\checkmark$ \\
    Gemma2\href{https://blog.google/technology/developers/google-gemma-2/}{$^{\triangle}$} & 9B, 27B & Multi. (UNK) + Code & Dec-Only & \cellcolor{blue!10}$\checkmark$ & \cellcolor{blue!10}$\checkmark$ \\
    Llama2~\citep{touvron2023llama2} & 7B, 13B, 70B & Multi. (UNK) + Code & Dec-Only & \cellcolor{blue!10}$\checkmark$ & \cellcolor{blue!10}$\checkmark$ \\
    Llama3, Llama3.1~\citep{dubey2024llama}{$^{\triangle}$} & 8B, 70B & Multi. (UNK) + Code & Dec-Only & \cellcolor{blue!10}$\checkmark$ & \cellcolor{blue!10}$\checkmark$ \\
    LlaMAX~\citep{lu2024llamax} & 7B, 8B & Multi. (102) & Dec-Only; Llama2, Llama3 & \cellcolor{blue!10}$\checkmark$ & $\times$\\
    Mistral~\citep{jiang2023mistral} & 7B & Multi. (UNK) + Code & Dec-Only & \cellcolor{blue!10}$\checkmark$ & \cellcolor{blue!10}$\checkmark$ \\
    Mixtral-MoE~\citep{jiang2024mixtral} & 8$\times$7B, 8$\times$22B & Multi. (UNK) + Code & MoE; Mistral & \cellcolor{blue!10}$\checkmark$ & \cellcolor{blue!10}$\checkmark$ \\ 
    Nemotron-4 (15B)~\citep{parmar2024nemotron} & 15B & Multi. (53) + Code (43) & Dec-Only & $\times$ & $\times$ \\
    Nemotron-4 (340B)~\citep{adler2024nemotron} & 340B & Multi. (53) + Code (43) & Dec-Only; Nemotron-4 (15B) & \cellcolor{blue!10}$\checkmark$ & \cellcolor{blue!10}$\checkmark$ \\
    NLLB~\citep{costa2022no} & 600M, 1.3B, 3.3B, 54.5B (MoE) & Multi. (200+) & Enc-Dec; M2M-100, MoE & \cellcolor{blue!10}$\checkmark$ & $\times$ \\
    Phi3~\citep{abdin2024phi} & 3.8B, 7B, 14B & Multi. (UNK) + Code & Dec-Only & \cellcolor{blue!10}$\checkmark$ & \cellcolor{blue!10}$\checkmark$  \\
    Qwen~\citep{bai2023qwen} & 1.8B, 7B, 14B, 72B & Multi. (100) + Code & Dec-Only & \cellcolor{blue!10}$\checkmark$ & \cellcolor{blue!10}$\checkmark$ \\
    Snowflake Artic\href{https://huggingface.co/Snowflake/snowflake-arctic-instruct}{$^{\triangle}$} & 128 $\times$ 3.66B & Multi. (UNK) + Code & MoE & \cellcolor{blue!10}$\checkmark$ & \cellcolor{blue!10}$\checkmark$ \\ 
    StableLM 2 (1.6B)~\citep{bellagente2024stable} & 1.6B & Multi. (7) + Code & Dec-Only & \cellcolor{blue!10}$\checkmark$ & \cellcolor{blue!10}$\checkmark$ \\
    StableVicuna\href{https://huggingface.co/CarperAI/stable-vicuna-13b-delta}{$^{\triangle}$} & 13B & English & Dec-Only; Vicuna & \cellcolor{blue!10}$\checkmark$ & \cellcolor{blue!10}$\checkmark$  \\
    Vicuna~\citep{chiang2023vicuna} & 7B, 13B & English & Dec-Only; Llama1, Llama2 & \cellcolor{blue!10}$\checkmark$ & $\times$  \\ \midrule
    \multicolumn{5}{l}{Close-weight and Close-source LM} \\ \midrule
    Bard~\citep{manyika2023overview} & UNK & UNK & UNK & \cellcolor{blue!10}$\checkmark$ & \cellcolor{blue!10}$\checkmark$ \\
    Chinchilla~\citep{hoffmann2022training} & 70B & English + Code & Dec-Only & $
    \times$ & $\times$\\
    Claude 3.5 Sonnet~\citep{anthropic2024claude} & UNK & UNK & UNK & \cellcolor{blue!10}$\checkmark$ & \cellcolor{blue!10}$\checkmark$ \\
    Command R (Plus)\href{https://cohere.com}{$^{\triangle}$} & UNK & UNK & UNK & \cellcolor{blue!10}$\checkmark$ & \cellcolor{blue!10}$\checkmark$ \\
    Gemini 1.0~\citep{team2023gemini} & UNK & UNK & Dec-Only & \cellcolor{blue!10}$\checkmark$ & \cellcolor{blue!10}$\checkmark$ \\
    Gemini 1.5~\citep{reid2024gemini} & UNK & UNK & MoE; Gemini 1.0 & \cellcolor{blue!10}$\checkmark$ & \cellcolor{blue!10}$\checkmark$ \\
    Gopher~\citep{rae2021scaling} & 280B & English + Code & Dec-Only & $\times$ & $\times$\\
    GPT-3~\citep{brown2020language} & 125M, ..., 175B & Multi. (UNK) & Dec-Only; GPT-2 & $\times$ & $\times$\\
    GPT-3.5 (Instruct GPT)~\citep{ouyang2022training} & 1.3B & UNK & Enc-Dec; GPT-3 & \cellcolor{blue!10}$\checkmark$ & \cellcolor{blue!10}$\checkmark$ \\
    GPT-4~\citep{achiam2023gpt} & UNK & Multi. (UNK) & UNK & \cellcolor{blue!10}$\checkmark$ & \cellcolor{blue!10}$\checkmark$ \\ 
    Reka~\citep{ormazabal2024reka} & 7B (Edge), 21B (Flash), UNK (Core) & Multi. (110) & Enc-Dec & \cellcolor{blue!10}$\checkmark$ & \cellcolor{blue!10}$\checkmark$
    \\ \midrule
    Close-access LM \\ \midrule
    AlexaTM~\citep{soltan2022alexatm} & 20B & Multi. (12) & Enc-Dec; BART & $\times$ & $\times$\\ 
    BloombergGPT~\citep{wu2023bloomberggpt} & 50.6B & English & Dec-Only; BLOOM & $\times$ & $\times$\\ 
    FLAN-PaLM~\citep{longpre2023flan} & 8B, 62B, 540B & Multi. (124+) + Code (24+) & UNK & \cellcolor{blue!10}$\checkmark$ & $\times$ \\
    PaLM~\citep{chowdhery2023palm} & 8B, 62B, 540B & Multi. (124) + Code (24) & Dec-Only & $\times$ & $\times$\\ 
    PaLM2~\citep{anil2023palm} &  400M, ..., 15B & Multi. (124+) + Code (24+) & UNK & \cellcolor{blue!10}$\checkmark$ & $\times$ \\
    \bottomrule
    \end{tabular}
}
\caption{Pre-trained Generative Language Models. $^\dagger$The languages do not include the languages seen by the base model.}
\label{tab:language-models}
\end{table*}

\begin{table*}[!t]
\centering
\resizebox{\textwidth}{!}{
    \begin{tabular}{llcccccc}
    \toprule
    \textbf{Model}  & \textbf{Sizes} & \textbf{SFT/Pref. Tuning Langs.$^\dagger$} & \multicolumn{1}{c}{\textbf{Model Base}} & \multicolumn{1}{c}{\textbf{SFT}} & \multicolumn{1}{c}{\textbf{Pref. Tuning}} \\ 
    \midrule
    \multicolumn{5}{l}{Open-weight SLM} \\ \midrule
    BAT~\citep{zheng2024bat} & 7B & English & Enc-Dec & \cellcolor{blue!10}$\checkmark$ & $\times$\\
    SpeechGPT~\citep{zhang2023speechgpt} & 13B & English & Dec & \cellcolor{blue!10}$\checkmark$ & \cellcolor{blue!10}$\checkmark$
    \\ \midrule
    \multicolumn{5}{l}{Open-source SLM} \\ \midrule
    \multicolumn{5}{l}{Close-weight and Close-source SLM} \\ \midrule Reka~\citep{ormazabal2024reka} & 7B (Edge), 21B (Flash), UNK (Core) & Multi. (110) & Enc-Dec & \cellcolor{blue!10}$\checkmark$ & \cellcolor{blue!10}$\checkmark$
    \\ \midrule
    \end{tabular}
}
\caption{Pre-trained Speech Language Models. $^\dagger$The languages do not include the languages seen by the base model.}
\label{tab:speech-language-models}
\end{table*}

\begin{table*}[!t]
\centering
\resizebox{0.98\textwidth}{!}{
    \begin{tabular}{llcccccc}
    \toprule
    \textbf{Model}  & \textbf{Sizes} & \textbf{SFT/Pref. Tuning Langs.$^\dagger$} & \multicolumn{1}{c}{\textbf{Model Base}} & \multicolumn{1}{c}{\textbf{SFT}} & \multicolumn{1}{c}{\textbf{Pref. Tuning}} \\ 
    \midrule
    \multicolumn{5}{l}{Open-weight VLM} \\ \midrule
    Falcon 2 VLM & 11B\href{https://huggingface.co/blog/falcon2-11b}{$^{\triangle}$} & Multi. (11) & Enc-Dec & \cellcolor{blue!10}$\checkmark$ & $\times$\\
    InstructBLIP~\citep{dai2023instructblip} & 7B, 13B (Vicuna) & English & Enc-Dec & \cellcolor{blue!10}$\checkmark$ & $\times$ \\
    & 3B, 11B (FLAN-T5) & English & Enc-Dec & \cellcolor{blue!10}$\checkmark$ & $\times$ \\
    InstructPix2Pix~\citep{brooks2023instructpix2pix} & UNK & English & UNK & \cellcolor{blue!10}$\checkmark$ & $\times$ \\
    LLaVA 1.5~\citep{liu2024improved} & 7B, 13B & English & Enc-Dec & \cellcolor{blue!10}$\checkmark$ & $\times$ \\
    LLaVA 1.6 (NeXT) & UNK\href{https://llava-vl.github.io/blog/2024-01-30-llava-next/}{$^{\triangle}$} & English & Enc-Dec & \cellcolor{blue!10}$\checkmark$ & $\times$ \\
    X-instructblip~\citep{panagopoulou2023x} & 7B, 13B & English & Enc-Dec & \cellcolor{blue!10}$\checkmark$ & $\times$ \\
    Phi3-Vision~\citep{abdin2024phi} & 4.2B & English & Enc-Dec & \cellcolor{blue!10}$\checkmark$ & $\times$ \\
    Otter~\citep{li2023mimic} & 7B (Dec) & English & Enc-Dec & \cellcolor{blue!10}$\checkmark$ & $\times$ \\
    MultiModal-GPT~\citep{gong2023multimodal} & UNK & English & Enc-Dec & \cellcolor{blue!10}$\checkmark$ & $\times$ \\ 
    Stable Diffusion v1.5~\citep{rombach2022high} & UNK & English & Enc-Dec & \cellcolor{blue!10}$\checkmark$ & $\times$ \\
    Video-LLaMA~\citep{zhang2023video} & 7B, 13B & English & Dec-Only & \cellcolor{blue!10}$\checkmark$ & $\times$ \\
    \midrule
    \multicolumn{5}{l}{Open-source VLM} \\ \midrule
    \multicolumn{5}{l}{Close-weight and Close-source SLM} \\ \midrule Reka~\citep{ormazabal2024reka} & 7B (Edge), 21B (Flash), UNK (Core) & Multi. (110) & Enc-Dec & \cellcolor{blue!10}$\checkmark$ & \cellcolor{blue!10}$\checkmark$ \\
    SORA~\citep{liu2024sora} & UNK & UNK & Enc-Dec & \cellcolor{blue!10}$\checkmark$ & \cellcolor{blue!10}$\checkmark$ \\ \midrule
    \end{tabular}
}
\caption{Pre-trained Vision Language Models. $^\dagger$The languages do not include the languages seen by the base model.}
\label{tab:vision-language-models}
\end{table*}

\subsection{Pre-trained Generative Models}
We categorize pre-trained generative models into three main types: LMs, VLMs, and SLMs. Additionally, we classify these models based on their accessibility: \textbf{(1) Open Source:} The model and data are open and accessible, \textbf{(2) Open-Weight:} Only the model is accessible and some or all data are inaccessible, \textbf{(3) Close-weight and Close-source:} The model is a black-box and may only be accessible by API or service, and \textbf{(4) Close Access:} The model is inaccessible. We also categorize these models based on the datasets used for pre-training, specifically noting whether they are trained with Supervised Fine-Tuning (SFT) datasets or Human Preference Tuning datasets.

\subsubsection{Language Models (LMs)}
Table~\ref{tab:language-models} shows the list of LMs categorized by the model accessibility and annotated with the model sizes, languages, model base, and fine-tuning methods applied to the model.

\subsubsection{Speech Language Models (SLMs)}
Table~\ref{tab:speech-language-models} shows the list of open-weight and open-source Speech Language Models (SLMs) categorized by the datasets and methods used in training.

\subsubsection{Vision Language Models (VLMs)}
Table~\ref{tab:vision-language-models} shows the list of open-weight and open-source Vision Language Models (VLMs) categorized by the datasets and methods used in training.

\section{Online Alignment}
In this section, we explore into human preference tuning using online methods, where data is continuously sampled. Online preference tuning involves real-time model updates as new data becomes available, enabling the model to dynamically adapt to evolving preferences and new information. This approach allows the alignment process to incorporate new data as it arrives and benefit from online exploration. We discuss the mechanisms of data collection, processing, and real-time model updates, emphasizing the benefits of managing non-stationary environments and enhancing model performance through continuous learning. Various techniques and strategies for implementing especially on-policy tuning are examined to provide a comprehensive understanding of its effective application in human preference tuning. We cover standard RL-based methods (e.g., PPO, which is online and on-policy), online DPO and SFT like algorithms (which can be on-policy or off-policy) and Nash Learning (or self-play) based algorithms.

\subsection{Reinforcement Learning Human Feedback (RLHF)}
In general, RLHF learns a reward function from human feedback and then optimize that reward
function~\citep{christiano2017deep}. The training for RLHF involves three stages:
\begin{itemize}
    \item The policy model $\pi_\theta$ interacts with the environment and the parameters of $\pi_\theta$ are updated via RL.
    \item The pairs of segments are selected from the output produced by the policy model $\pi_\theta$, and send them to human annotators for comparison.
    \item The parameters are optimized using reward $r$ to fit the comparisons collected from human.
\end{itemize}
According to \cite{ziegler2019fine}, the RLHF pipeline for LMs can be summarized as following:
\begin{itemize}
    \item \textbf{Supervised Fine-Tuning:} A pre-trained LM is instruction-tuned using a dataset consisting of a given instruction prompt, and (typically) a human-written completion. The LM/policy is trained with a cross-entropy loss over the completion only. Often, the SFT model, denoted as sft is used to initialize both the reward model and the RLHF policy.
    \item \textbf{Reward Modeling:} RLHF leverages a reward model $r_\phi$ trained using a dataset of preferences $\mathcal{D}$. The reward model is trained using the following loss:
    \begin{equation}
    \label{eq:rploss}
    \textrm{loss}(r) = \E{\left(x, \left\{y_i\right\}_i, b\right) \sim S}{\log {\frac {e^{r(x, y_b)}}{\sum_i e^{r(x, y_i)}}}}.
    \end{equation}
    or, for pairwise preferences,
    \begin{align}
    \mathcal{L}_{\text{RM}}(\phi) = -\mathbb{E}_{(x, y_w, y_l)\sim\mathcal{D}_\text{pref}}\log \sigma(r_{\phi}(x,y_w)-r_{\phi}(x,y_l)).
    \end{align}
    \item \textbf{Reinforcement Learning:} In this stage, the learned reward model $r_{\phi^*}$ is used to provide online feedback in the optimization of the policy. In \cite{ziegler2019fine,stiennon2020learning,ouyang2022training}, RLHF further maximizes average reward with an extra KL regularization term, i.e.:
    \begin{align}
    \label{RLHF objective}
    \mathcal{L}_{\text{RL}}(\phi) = \mathbb{E}_{x\sim \mathcal{D}, y\sim{\pi(\cdot\mid x)}}\left[r_{\phi^*}(x,y)-\beta_\text{reg} \operatorname{KL}(\pi(\cdot\mid x)\mid \pi_{\text{ref}}(\cdot\mid x))\right],
    \end{align}
    where $\beta_\text{reg} > 0$ is a hyper-parameter controlling the deviation from the reference policy
    $\pi_{\mathrm{ref}} = \pi^{\mathrm{SFT}}$.
\end{itemize}

RLHF proposes optimizing the policy model using the Advantage Actor-Critic (A2C) method~\citep{mnih2016asynchronous} for playing Atari games and Trust Region Policy Optimization (TRPO)~\citep{mnih2015human} for performing simulated robotics tasks. The reward model is trained using the Bradley-Terry Reward Model, which leverages pairwise preference datasets—essentially, pairs of preferred and dispreferred responses. There are various methods and variations for training RLHF, primarily categorized into two main approaches: RLHF and REINFORCE. In the following sections, we will describe these methods in detail.

\subsubsection{Proximal Policy Optimization (PPO)}
Initially in the original RLHF paper~\citep{ziegler2019fine}, they use PPO~\citep{schulman2017proximal} as their optimization strategy. PPO framework is a method for the human preference signals from external reward models with RLHF. The idea is to improve the current state of affairs by introducing an algorithm that attains the data efficiency and reliable performance of TRPO, while using only first-order optimization with a simpler clipped surrogate objective, omitting the expensive second-order optimization presented in TRPO using stochastic gradient ascent. Whereas standard policy gradient methods perform one gradient update per data sample, PPO~\citep{schulman2017proximal} proposes a novel objective function that enables multiple epochs of minibatch updates. It have some of the benefits of TRPO, but they are much simpler to implement and more efficient. For the optimization, KL-shaped reward~\citep{ahmadian2024back} is useful as penalty-free optimization of the reward model leads to degradation in the coherence of the model. Optimizing this objective is equivalent to maximizing the following KL-shaped reward in expectation. There are a couple of variants of PPO: A2C~\citep{mnih2016asynchronous}, P3O~\citep{wu2023pairwise}, PTR-PPO~\citep{liang2021ptr}, and RLHF-V~\citep{yu2024rlhf}.

\paragraph{Advantage Actor-Critic (A2C)}
A2C~\citep{mnih2016asynchronous} is an asynchronous variant of four RL algorithms that utilize parallel actor-learners to stabilize the effect of training of four methods. 

\paragraph{Pairwise Proximal Policy Optimization (P3O)} 
P3O~\citep{wu2023pairwise} is an on-policy RL algorithms that interleaves off-policy updates with on-policy updates. P3O uses the effective sample size between the behavior policy and the target policy to control how far they can be from each other and does not introduce any additional hyper-parameters.

\paragraph{Prioritized Trajectory Replay (PTR-PPO)}
PTR-PPO~\citep{liang2021ptr} is an on-policy deep reinforcement learning algorithms have low data utilization and require significant experience for policy improvement. The algorithm proposes a proximal policy optimization algorithm with PTR-PPO that combines on-policy and off-policy methods to improve sampling efficiency by prioritizing the replay of trajectories generated by old policies. The method is designed three trajectory priorities based on the characteristics of trajectories: the first two being max and mean trajectory priorities based on one-step empirical generalized advantage estimation (GAE) values and the last being reward trajectory priorities based on normalized undiscounted cumulative reward. Then, it is also incorporated the prioritized trajectory replay into the PPO algorithm, propose a truncated importance weight method to overcome the high variance caused by large importance weights under multistep experience, and design a policy improvement loss function for PPO under off-policy conditions.

\paragraph{RLHF-V}
RLHF-V~\citep{yu2024rlhf} enhances MLLM trustworthiness via behavior alignment from fine-grained correctional human feedback. Specifically, RLHF-V collects human preference in the form of segment-level corrections on hallucinations, and performs dense direct preference optimization over the human feedback. 

\subsubsection{REINFORCE}
\paragraph{ReMax}
ReMax~\citep{li2023remax} builds upon the well-known REINFORCE algorithm \citep{williams1987reinforcement, williams1992simple}, leveraging three key properties of RLHF: fast simulation, deterministic transitions, and trajectory-level rewards. The name ``ReMax" reflects its foundation in REINFORCE and its use of the argmax operator. ReMax modifies the gradient estimation by incorporating a subtractive baseline value as following:
\begin{align}
    \widetilde{g}(\theta) = \frac{1}{N} \sum_{i=1}^{N} \sum_{t=1}^{T} \big[s_{\theta}(x^{i},a_{1:t}^{i}) \times (r(x^{i}, a_{1:T}^{i}) - b_{\theta}({x^{i}}) ) \big], 
\end{align}
where the action $a_t^{i} \sim \pi_{\theta}(\cdot|x^{i}, a_{1:t-1}^{i})$, and $b_{\theta}(x^{i})$ is a baseline value. A typical choice for $b_{\theta}(x^{i})$ is 
\begin{align}
    b_{\theta}(x^{i}) = r(x^{i}, \bar{a}_{1:T}^{i}), \, \bar{a}_{t}^{i} \in \argmax  \pi_{\theta}(\cdot|x^{i}, \bar{a}^{i}_{1:t-1}).
\end{align}
This baseline value can be obtained by greedily sampling a response and calculating the associated reward value.

\paragraph{REINFORCE Leave One-Out (RLOO)}
RLOO~\citep{ahmadian2024back} extends the REINFORCE algorithm by leveraging multiple online samples to achieve unbiased variance reduction. It improves upon REINFORCE in two key ways: (1) The rewards from each sample can serve as a baseline for all other samples, and (2) Policy updates are performed using the average of gradient estimates from each sample, resulting in a variance-reduced multi-sample Monte Carlo (MC) estimate. This is the intuition behind the RLOO estimator, as following: 
\begin{align}
    \frac{1}{k}&\sum_{i=1}^k [R(y_{(i)},x) - \frac{1}{k-1}\sum_{j\ne{i}}R(y_{(j)},x)] \nabla \log \pi(y_{(i)}|x), \text{      for   } y_{(1)},...,y_{(k)} \overset{i.i.d}{\sim} \pi_\theta (.|x),    
    \label{eq:RLOO}
\end{align}
where $k$ refers to the number of online samples generated, $\textsc{RLOO}_{k}$ considers each $y_{(i)}$ individually and uses the remaining $k-1$ samples to create an unbiased estimate of the expected return for the prompt. This approach functions similarly to a parameter-free value function, but it is estimated at each training step.

\subsection{Online Directed Preference Optimization (Online DPO)}
\label{subsec:online dpo}
\subsubsection{Online AI Feedback (OAIF)}
OAIF~\citep{guo2024direct} employs a LLM as an annotator during each training iteration. In this process, two responses are sampled from the current model, and the LLM annotator is prompted to select the preferred response, thereby providing real-time feedback. OAIF aims to gather preferences dynamically for responses generated by the language model being aligned. Given the prohibitive cost of using human feedback, this method leverages an LLM as an online annotator to collect preferences over pairs of responses sampled from the model $\pi_\theta$ during its alignment process. The objective for online DPO yields (please see detailed derivation of DPO in Section \ref{subsec:DPO}):
\begin{multline}
\label{online DPO objective}
\mathcal{L}_{\mathrm{OAIF}}\left(\pi_{\theta} ; \pi_{\mathrm{ref}}\right)
:=\\
-\mathbb{E}_{x\sim\mathcal{D},\left(y_w, y_l\right) \sim \pi_{\theta_{-}}}\left[\log \sigma\left(\beta_\text{reg} \log \frac{\pi_{\theta} \left(y_w \mid x\right)}{\pi_{\text {ref }}\left(y_w \mid x\right)}-\beta_\text{reg} \log \frac{\pi_{\theta} \left(y_l \mid x\right)}{\pi_{\text {ref }}\left(y_l \mid x\right)}\right)\right],
\end{multline}
in which we note $\pi_{\theta_{-}}$ to show that preference pairs are generated under $\pi_{\theta}$, but we further adopt a stop gradient to prevent it from getting into the loss objective for the gradient computation. The OAIF is illustrated in Algorithm \ref{alg:OAIF} (OAIF algorithm in \cite{guo2024direct}), in which function $\ell$ can be log-sigmoid (DPO), square (IPO), or ReLU (SLiC) functions.

\begin{algorithm}[ht]
\caption{Online AI Feedback (OAIF) for Direct Alignment from Preference (DAP)}
\label{alg:OAIF}
\begin{algorithmic}[1]
 \State \textbf{Input:} Prompt dataset $\mathcal{D}_x = \{x_i\}_{i=1}^{N}$, an LLM annotator, SFT model $\pi_{\theta^0}$
 \For{$t := 0$ to $T$}
    \State Sample prompt $x \sim \mathcal{D}_x$
    \State Sample response pair $y_1, y_2 \sim \pi_{\theta^t}(\cdot|x)$
    \State Use LLM annotator to get preference pair $y_w$, $y_l$
    \State Update $\theta^t$ into $\theta^{t+1}$ using $\nabla_\theta \ell(x, y_w, y_l, \theta^t)$
 \EndFor
\end{algorithmic}
\end{algorithm}

\subsubsection{Iterative Directed Preference Optimization}
Iterative DPO \citep{xu2023some,xiong2024iterative} has been proposed to narrow the gap between the performance offline preference optimization methods like DPO and online methods like RLHF, as RLHF still outperforms offline DPO. Different from DPO that used a fixed offline dataset, iterative DPO proposed to formulate the preference datasets by the generations of the current model and labelers, being either a pretrained reward model or LLM as a judge or the model to be trained itself through specific prompting \citep{yuan2024self}, thus this pipeline usually appears at the same time with self-rewarding \citep{yuan2024self} methods (some paper will even call self-rewarding as iterative DPO methods). For each iteration, if the batch size for preference datasets utilized for policy optimization is only 1, then iterative DPO is essentially the same as online DPO or OAIF, except that the reference policy may be chosen as the last iterated policy instead of always being the SFT policy; otherwise iterative DPO is a hybrid method which combines offline learning in loss function optimization and online sampling in preference data generation. The reference model in the loss objective may differ between different methods, can be fixed SFT model \cite{xiong2024iterative} or last iterated model \citep{xu2023some,yuan2024self} or some mixtures.

\subsubsection{Online Preference Tuning (OPTune)}
OPTune~\citep{chen2024optune} is an algorithm for efficient data generation in online RLHF. It improves both generation and training efficiency by selectively regenerating only the lowest-rewarded responses and employing a weighted DPO objective that prioritizes pairs with larger reward gaps. This approach significantly enhances the overall efficiency of the RLHF pipeline, setting the stage for the development of preference-aligned LLMs in a resource-efficient manner. The method enhances both data generation and training efficiency for online preference alignment. To minimize the cost of iterative data regeneration, it employs a straightforward yet effective reward-based prompt selection strategy, updating responses only for prompts with the lowest scores according to the reward model. Additionally, recognizing that converting scalar rewards to binary labels for the online DPO objective results in information loss, the method introduces a weighted DPO loss variant. This variant prioritizes learning from response pairs with larger reward gaps, further boosting online learning efficiency. 

\subsection{SFT-like}
\subsubsection{Rank Responses to align Human Feedback (RRHF)}
RRHF~\citep{yuan2023rrhf} is a method that evaluates sampled responses from various sources using the logarithm of conditional probabilities and aligns these probabilities with human preferences through ranking loss. This approach can utilize responses from multiple origins, including the model's own outputs, responses from other large language models, and human expert responses, to learn how to rank them effectively. The primary objective is to simplify the complex hyper-parameter tuning and extensive training resources required by PPO. Before training, RRHF samples responses from diverse sources, which can include model-generated responses from the model itself as well as pre-existing human-authored responses of varying quality. During training, RRHF scores these responses based on the log probability provided by the training language model. These scores are then aligned with human preference rankings or labels using ranking loss, ensuring that the model's outputs are better aligned with human preferences.

\subsubsection{Reward rAnked FineTuning (RAFT)} 
RAFT~\citep{dong2023raft} is the combination of ranking samples by rewards and SFT, which iteratively alternates among three steps: 1) The batch is sampled from the generative models; 2) The reward function is used to score the samples and filter them to get a filtered subset of high rewards; and 3) fine-tune the generative models on the filtered subset.

\subsubsection{Reinforced Self-Training (ReST)}
ReST~\citep{gulcehre2023reinforced} is an RLHF algorithm aimed at aligning an LM's outputs with human preferences. It uses a learned reward function to model human preferences over sequences. In the Markov decision process underlying conditional language modeling, states represent partial sequences, and actions correspond to generated tokens. ReST divides the typical reinforcement learning pipeline into distinct offline stages for dataset growth and policy improvement. Initially, it fine-tunes a model to map input sequences to output sequences using a dataset of sequence pairs, optimizing with Negative Log-Likelihood (NLL) loss. Then, it creates a new dataset by augmenting the initial training dataset with samples generated by the model. In this phase, conditioning inputs are resampled from the original dataset, similar to self-training, but direct sampling is possible if accessible.

\subsubsection{Supervised Iterative Learning from Human Feedback (SuperHF)}
SuperHF~\citep{mukobi2023superhf} is an alignment algorithm that enhances data efficiency using a reward model and replaces PPO with a straightforward supervised fine-tuning loss. The core concept involves the language model generating its own training data by sampling a ``superbatch" of outputs, filtering these through a reward model, and iteratively fine-tuning on each filtered completion. This method builds upon and unifies previous research by integrating two crucial components: (1) the Kullback-Leibler (KL) divergence penalty and (2) an iterative process of sampling and fine-tuning. Additionally, SuperHF is embedded within a Bayesian inference framework, demonstrating that both RLHF and SuperHF can be understood from a unified theoretical perspective that does not rely on reinforcement learning. This perspective naturally justifies the use of the KL penalty and the iterative approach.

\subsection{Nash Learning}
\subsubsection{Nash Learning from Human Feedback (NLHF)}
NLHF~\citep{munos2023nash} is motivated to address the limitation of reward models (or essentially the Elo ratings) to represent the richness of human preferences as in RLHF. Instead of targeting at maximizing the (regularized) reward, NLHF takes the preference model as the `first class citizen', and pursue `a policy that consistently generates responses preferred
over those generated by any competing policy'. Thus this policy is the Nash equilibrium of this preference
model, the reason the method is named NLHF. Concretely, the (regularized) preference model for two policies $\pi,\pi^\prime$ is defined as:
\begin{multline}
\mathcal{P}\left(\pi>\pi^{\prime}\right) := \\
\mathbb{E}_{x \sim \rho} \mathbb{E}_{y \sim \pi(\cdot \mid x), y^{\prime} \sim \pi^{\prime}(\cdot \mid x)}\left[\mathcal{P}\left(y>y^{\prime} \mid x\right)-\beta_{\text{reg}} \log \frac{\pi(y \mid x)}{\mu(y \mid x)}+\beta_{\text{reg}} \log \frac{\pi^{\prime}\left(y^{\prime} \mid x\right)}{\mu\left(y^{\prime} \mid x\right)}\right],
\end{multline}
and NLHF searches the Nash Equilibrium such that (denote $\mu$ as $\pi_{\text{ref}}$ for simplicity here):
\begin{equation}
\pi^* := \arg \max _\pi \min _{\pi^{\prime}} \mathcal{P}\left(\pi>\pi^{\prime}\right)-\beta_{\text{reg}}\mathrm{KL}_\rho(\pi, \mu)+\beta_{\text{reg}}\mathrm{KL}_\rho\left(\pi^{\prime}, \mu\right).
\end{equation}
For optimization, the Nash-MD algorithm proposed in NLHF used a geometric mixture between the current policy $\pi_t$ and the reference policy $\mu$ as the competing policy in the place of $\pi^\prime$:
\begin{equation}
\pi_t^\mu(y) := \frac{\pi_t(y)^{1-\eta \beta_{\text{reg}}} \mu(y)^{\eta \beta_{\text{reg}}}}{\sum_{y^{\prime}} \pi_t\left(y^{\prime}\right)^{1-\eta \beta_{\text{reg}}} \mu\left(y^{\prime}\right)^{\eta \beta_{\text{reg}}}},
\end{equation}
where $\eta$ is a learning rate, and Nash-MD algorithm is a step of mirror descent relative to the regularized policy $\pi_t^\mu$:
\begin{equation}
\pi_{t+1} := \arg \max _\pi\left[\eta \mathcal{P}\left(\pi>\pi_t^\mu\right)-\operatorname{KL}\left(\pi, \pi_t^\mu\right)\right],
\end{equation}
which yields a closed-form solution that:
\begin{equation}
\log \pi_{t+1}(y)=\left[(1-\eta \beta_{\text{reg}}) \log \pi_t(y)+\eta \beta_{\text{reg}} \log \mu(y)\right]+\eta \mathcal{P}\left(y>\pi_t^\mu\right)+c,
\end{equation}
where $c$ is a normalization constant which is independent of $y$ and the algorithm is proved to converge of rate $\frac{1}{T}$ under the tabular setting.
For practical concern, when policy is a deep neural network beyond tabular setting, NLHF further proposes Nash-MD-PG motivated by Nash-MD, and the algorithm updates the policy with policy gradient:
\begin{equation}
\begin{aligned}
&\nabla_\theta \mathcal{P}_\tau\left(\pi_\theta>\pi^{\prime}_{\theta_-}\right)=\mathbb{E}_{x \sim \rho,
y \sim \pi_\theta(\cdot \mid x),
y^{\prime} \sim \pi^{\prime}(\cdot \mid x)}\left[\widehat{g}\left(x, y, y^{\prime}\right)\right],
\end{aligned}
\end{equation}
where  $\pi^{\prime}_{\theta_-}$ denotes a stop-gradient on $\pi^{\prime}_{\theta}$ with $\pi^{\prime}_{\theta}$ being a geometric mixture
\begin{equation}
\log \pi^{\prime}_{\theta}(y \mid x) := (1-\lambda) \log \left(\pi_\theta(y \mid x)\right)+\lambda \log (\mu(y \mid x))+c(x),
\end{equation}
in which $\lambda$ is a mixing constant and
\begin{equation}
\widehat{g}\left(x, y, y^{\prime}\right) := \nabla_\theta \log \pi_\theta(y \mid x)\left(\mathcal{P}\left(y>y^{\prime} \mid x\right)-1 / 2-\beta_{\text{reg}} \operatorname{KL}\left(\pi_\theta(\cdot \mid x), \mu(\cdot \mid x)\right)\right),
\end{equation}
respectively. NLHF also argues that, Nash equilibrium of the preference model is a solution that better aligns with the diversity of human preferences.

\subsubsection{Self-Play Preference Optimization (SPPO)}
SPPO \citep{wu2024self} can be understood as a specific instance of NLHF by taking $\lambda=0$, i.e., the reference policy is itself. The algorithm can be found in Algorithm \ref{alg:sppo}, given an LLM judge:
\begin{algorithm}[ht]
\caption{Self-Play Preference Optimization (SPPO)}
\label{alg:sppo}
\begin{algorithmic}[1]
 \State \textbf{Input:} base policy $\pi_{\theta_0}$, preference oracle $\mathcal{P}$, learning rate $\eta$, number of generated samples $K$
 \For{$t = 0, 1, \dots$}
    \State Generate synthetic responses by sampling $x \sim \mathcal{D}$ and $y_{1:K} \sim \pi_{\theta_t}(\cdot|x)$
    \State Annotate the win-rate $\mathcal{P}(y_k \succ y_{k'}|x), \forall k, k' \in [K]$
    \State Select responses from $y_{1:K}$ to form dataset $D_t = \{(x_i, y_i, \hat{\mathcal{P}}(y_i \succ \pi_{\theta_t}|x_i))\}_{i \in [N]}$
    \State Optimize $\pi_{\theta_{t+1}}$ according to:
    \[
    \theta_{t+1} \leftarrow \arg\min_{\theta} \mathbb{E}_{(x, y, \hat{\mathcal{P}}(y \succ \pi_{\theta_t}|x)) \sim D_t} \left( \log \left( \frac{\pi_{\theta}(y|x)}{\pi_{\theta_t}(y|x)} \right) - \eta \left( \hat{\mathcal{P}}(y \succ \pi_{\theta_t}|x) - \frac{1}{2} \right)^2 \right).
    \]
 \EndFor
\end{algorithmic}
\end{algorithm}

\subsection{Fine-tuning Diffusion Models}
Given the popularity of diffusion based t2I models and its different nature of structural properties, we have the methods of fine-tuning diffusion models as a separate section of interest. We first briefly review the formulation of text-to-image diffusion generative models. For a more comprehensive  background of diffusion models, we refer the interested readers to existing tutorial/survey papers~\citep{luo2022understanding,cao2024survey,yang2023diffusion,tang2024score,chen2024overview,chan2024tutorial}. DDPM \citep{sohl2015deep,Ho20DDPM} consider a sequence of positive noise scales $0<\beta_1, \beta_2, \cdots, \beta_N<1$, and perturb data by gradually adding noise through a stochastic process: for each training data point $x_0 \sim p_{\text {data }}(x)$, a discrete Markov chain $\left\{x_0, x_1, \cdots, x_N\right\}$ is constructed such that:
\begin{equation}
\label{DDPM forward}
x_i=\sqrt{1-\beta_i} x_{i-1}+\sqrt{\beta_i} z_{i-1}, \quad i=1, \cdots, N,
\end{equation}
where $z_{i-1} \sim \mathcal{N}(0, I)$. For generative modeling, the backward process - a variational Markov chain in the reverse direction - is parameterized with 
\begin{equation}
p_{\theta}\left(x_{i-1} \mid x_i\right)=\mathcal{N}\left(x_{i-1} ; \frac{1}{\sqrt{1-\beta_i}}\left(x_i+\beta_i s_{\theta}\left(i,x_i\right)\right), \beta_i I\right),
\end{equation}
in which $s_{\theta}\left(i,x_i\right)$ is learned by maximizing an evidience lower bound (ELBO). In the context of text-to-image generation, trained $s_{\theta^*}$ will also be dependent on an input prompt $c$ for conditional generation. For inference, samples can be generated by starting from pure noise and following the estimated reverse process as:
\begin{equation}
\label{DDPM Backward Process}
x_{i-1}=\frac{1}{\sqrt{1-\beta_i}}\left(x_i+\beta_i s_{\theta^*}\left(i,x_i,c\right)\right)+\sqrt{\beta_i} z_i, \quad i=N, N-1, \cdots, 1 .
\end{equation}

\subsubsection{DDPO and DPOK}
We review some key elements in DDPO and DPOK \citep{black2024training,fan2024reinforcement} to formulate the problem of fine-tuning diffusion models as discrete-time MDPs, and then apply RL algorithms. Note that recent works, \cite{tang2024fine,uehara2024understanding,uehara2024fine} extend a continuous-time formulation for fine-tuning, but we stick to the discrete time case for simplicity. Consider taking $(i,x_i,c)$ as the state space, and define the action as the next hierarchy $x_{i-1}$ to go to, then Eq. \eqref{DDPM Backward Process} naturally defines a stochastic policy: the stochasticity of the policy comes from $\sqrt{\beta_{i}} z_i$, thus the policy follows Gaussian with mean determined by $s_{\theta^*}\left(i,x_i,c\right)$ with variance $\beta_i$:
\begin{equation}
    \pi_{\theta}(x_{i-1}\mid x_{i})\sim\mathcal{N}\left(\frac{1}{\sqrt{1-\beta_i}}\left(x_i+\beta_i s_{\theta}\left(i,x_i,c\right)\right),\beta_i\right), \quad i=N, N-1, \cdots, 1 .
\end{equation}
Given this formulation, \cite{black2024training} directly maximize the expected reward (without regularization) $\mathcal{J}_{\text {DDPO}}= \mathbb{E}_{\theta}\left[r(x_0,c)\right]$ by REINFORCE or PPO:
\begin{equation}
\nabla_\theta \mathcal{J}_{\text {DDPO }}=\mathbb{E}\left[\sum_{t=0}^T \nabla_\theta \log p_\theta\left(x_{t-1} \mid x_t, c\right) r\left(x_0, c\right)\right].
\end{equation}
Compare to DDPO, DPOK~\citep{fan2024reinforcement} optimize the same regularized reward objective as in Eq. \eqref{RLHF objective}:
\begin{equation}
\mathcal{J}_{\text {DPOK }}= \mathbb{E}_{\theta}\left[r(x_0,c)\right]-\beta \mathbb{E}_{p(z)}\left[\operatorname{KL}\left(p_\theta\left(x_0 \mid z\right) \|\, p_{\text {pre }}\left(x_0 \mid z\right)\right)\right]
\end{equation}
They further proposed a clipped gradient algorithm for optimization, motivated by the original PPO clipped objective. In addition, DPOK shows that adding regularization will yield a better generation result compared to the version without regularization.

\subsubsection{Reward Feedback Learning (ReFL)}
ReFL~\citep{xu2024imagereward} is a supervised fine-tuning method based on its pre-trained reward model ImageReward $r_{\text{IR}}(c,x)$. The objective for ReFL optimization is a linear combination of negative pre-trained loss (for diffusion models) and reward maximization:
\begin{equation}
\mathcal{J}_{\text{ReFL}}(\theta) =\mathcal{J}_{\text{pre}}(\theta)+\lambda ~\mathbb{E}_{c \sim p_{\boldsymbol{c}},x_t\sim p_{\theta}(\cdot\mid c)}\left(\phi\left(r_{\text{IR}}\left(c, x_t\right)\right)\right),
\end{equation}
in which $\lambda$ is a scaling constant, $\phi$ is taken as a ReLU function and $t\in[0,\tilde{T}]$ is a random number for sampling, a technique that \citep{xu2024imagereward} claims can help stabilize the training instead of always letting $t$ be 0.

\subsubsection{Direct Reward Fine-Tuning  (DRaFT)}
DRaFT~\citep{clark2023directly} introduces a straightforward method for fine-tuning diffusion models using differentiable reward functions. The goal is to fine-tune the parameters $\theta$ of a pre-trained diffusion model such that images generated by the sampling process maximize a differentiable reward function $r$: 
\begin{align} 
\label{eq:reward-objective}
J(\theta) = \mathbb{E}_{\boldsymbol{c} \sim p_{\boldsymbol{c}}, \boldsymbol{x}_T \sim \mathcal{N}(\boldsymbol{0}, \boldsymbol{I)}} \left[ r(\text{sample}(\theta, \boldsymbol{c}, \boldsymbol{x}_T), \boldsymbol{c}) \right],
\end{align}
where $\text{sample}(\theta, \boldsymbol{c}, \boldsymbol{x}_T)$ denotes the sampling process from time $t = T \to 0$ with context $\boldsymbol{c}$. First, DRaFT consider solving Eq.~\ref{eq:reward-objective} by computing $\nabla_{\theta} r(\text{sample}(\theta, \boldsymbol{c}, \boldsymbol{x}_T), \boldsymbol{c})$ and using gradient ascent. Computing this gradient requires backpropagation through multiple diffusion model calls in the sampling chain, similar to backpropagation through time in a recurrent neural network. To mitigate the memory cost associated with this process, DRaFT employs two strategies: 1) low-rank adaptation (LoRA) \citep{hu2021lora}, and 2) gradient checkpointing \citep{chen2016training}.

\subsubsection{AlignProp}
AlignProp~\citep{prabhudesai2023aligning} introduces a method that transforms denoising inference within text-to-image diffusion models into a differentiable recurrent policy, effectively linking conditioning input prompts and sampled noise to generate output images. This approach enables fine-tuning of the denoising model's weights through end-to-end backpropagation, guided by differentiable reward functions applied to the generated images. The proposed model casts conditional image denoising as a single step MDP with states $\mathcal{S}= \{(x_T,\textbf{c} ), x_T\sim\mathcal{N}(\mathbf{0},\mathbf{I})\}$,   actions are the generated image samples, and the whole DDIM denoising chain corresponds to a differentiable policy that maps states to image samples: $\mathcal{A} = \{ x_0: x_0 \sim \pi_\theta(\cdot | x_T,\textbf{c}), x_T\sim\mathcal{N}(\mathbf{0},\mathbf{I})\ \} $. The reward function is a differentiable function of parameters $\phi$  that depends only on generated images  $R_\phi(x_0), x_0 \in \mathcal{A}$. Given a dataset of prompts input $\mathcal{D}$, our loss function reads:
\begin{equation}
\label{eq:loss}
\mathcal{L}_\text{align}(\theta;\mathcal{D})
= - 
\tfrac{1}{|\mathcal{D}|}
\sum_{\textbf{c}^i\in\mathcal{D}}  R_\phi(\pi_{\theta}(x_T, \textbf{c}^i)).
\end{equation}
The parameters of the diffusion model using gradient descent on $\mathcal{L}_\text{align}$. The policy $\pi$ is recurrent, and training it is akin to backpropagation through time, a technique commonly used for training recurrent neural networks. The gradient for updating the parameters of the diffusion model with respect to the downstream objective (i.e., the differentiable reward function) is expressed as following:
\begin{align}
\label{eq:single_backprop}
\hat{\nabla}_\theta \mathcal{L}_\text{align} = \frac{\partial \mathcal{L}_\text{align}}{\partial \theta} + \sum_{t = 0}^K \frac{\partial\mathcal{L}_\text{align}}{\partial x_{t}} \cdot \frac{\partial x_{t}}{\partial \theta}, 
\end{align}
in which $K$ is uniformly drawn from $[0,T]$ for memory efficiency instead of being $T$, referred as randomized truncation in~\cite{prabhudesai2023aligning}.

\subsubsection{Proximal Reward Difference Prediction}
PRDP \citep{deng2024prdp} proposed a loss for matching likelihood difference with reward difference for fine-tuning diffusion models, inspired by DPO. Notice that, (same as derivation in DPO), for any two generations $x^1_0$ and $x^2_0$, the optimal policy (KL-regularized reward) yields:
\begin{equation}
\log \frac{\pi_{\theta^{\star}}\left(x^1_0 \mid \mathbf{c}\right)}{\pi_{\text {ref }}\left(x^1_0 \mid \mathbf{c}\right)}-\log \frac{\pi_{\theta^{\star}}\left(x^2_0\mid \mathbf{c}\right)}{\pi_{\text {ref }}\left(x^2_0 \mid \mathbf{c}\right)}=\frac{r\left(x_0^1, \mathbf{c}\right)-r\left(x_0^2, \mathbf{c}\right)}{\beta_{\text{reg}}}
\end{equation}
thus PRDP proposes to minimize the MSE error between LHS with $\theta$ (replacing $\theta^*$) and RHS. The objective is $\mathcal{L}_{\mathrm{PRDP}}\left(\pi_{\theta} ; \pi_{\mathrm{ref}}\right)
:=$ 
\begin{align}
\label{PRDP objective}
\mathbb{E}_{c\sim\mathcal{D},\left(x^1, x^2\right) \sim \pi_{\theta}(\cdot\mid c)}\left(\beta_\text{reg} \log \frac{\pi_{\theta} \left(x_0^1 \mid x\right)}{\pi_{\text {ref }}\left(x_0^1 \mid x\right)}-\beta_\text{reg} \log \frac{\pi_{\theta} \left(x_0^2 \mid x\right)}{\pi_{\text {ref }}\left(x_0^2 \mid x\right)}-(r(x^1_0)-r(x^2_0))\right)^2,
\end{align}
Furthermore, they also employ proximal updates (clipping the ratios and optimizing a proximal objective) for stable training of \eqref{PRDP objective}, in the same spirit of PPO. 

Similar works include \cite{yang2024dense}, which applies the idea of dense reward to DPO-style explicit-reward-free approach on text-to-image diffusion models, so as to suit better to diffusion models' generation hierarchy. 

\subsubsection{Diffusion Loss-guided Policy Optimization (DLPO)}
DLPO~\citep{chen2024reinforcement} applies online RL to fine-tune TTS diffusion models, where the reward is shaped by the diffusion model’s loss. Incorporating the diffusion model loss into the objective function serves as an additional mechanism to enhance performance and maintain the coherence of the model. The method's objective is described as following:
\begin{equation}
\label{eq10}
    \mathbb{E}_{c \sim p(c)}\mathbb{E}_{t\sim \mathcal{U}\{1,T\}}\mathbb{E}_{p_\theta(x_{0:T}|c)} \left[-\alpha r(x_0,c)-\beta\Vert \tilde{\epsilon}(x_t,t) - \epsilon_\theta(x_t,c,t)\Vert_2 \right],
\end{equation}
where $\alpha,\beta$ are the reward and weights for diffusion model loss, respectively. DLPO uses the following gradient to update the objective:
\begin{multline}
\label{eq11}
    \mathbb{E}_{c\sim p(c)}\mathbb{E}_{t\sim \mathcal{U}\{0,T\}} \mathbb{E}_{p_\theta(x_{1:T}|c)}\\
    \left[- \left(\alpha r(x_0,c)-\beta\nabla_\theta\Vert \tilde{\epsilon}(x_t,t) - \epsilon_\theta(x_t,c,t)\Vert_2\right) \nabla_\theta \log p_\theta(x_{t-1}|x_t,c)\right].
\end{multline}
The diffusion model objective is incorporated into the reward function as a penalty. This algorithm aligns with the training procedure of TTS diffusion models by integrating the original diffusion model objective $\beta\Vert \tilde{\epsilon}(x_t,t) - \epsilon_\theta(x_t,c,t)\Vert_2 $ as a penalty in the reward function. This approach effectively prevents model deviation and ensures that the model remains coherent during training.

\subsubsection{Human Feedback for Instructional Visual Editing (HIVE)}
HIVE~\citep{zhang2024hive} is proposed to improve instruction visual editing models (diffusion models based, e.g., InstructPix2Pix~\citep{brooks2023instructpix2pix}) with human feedback. In instructional supervised training, the stable diffusion model has two conditions $c=\left[c_I, c_T\right]$, where $c_T$ is the editing instruction, and $c_I$ is the latent space of the original input image. In the training process, a pre-trained auto-encoder with encoder $\mathcal{E}$ and decoder $\mathcal{D}$ is used to convert between edited image $\tilde{x}$ and its latent representation $z=\mathcal{E}(\tilde{x})$. The diffusion process is composed of an equally weighted sequence of denoising autoencoders $\epsilon_\theta\left(z_t, t, c\right)$, $t=1, \cdots, T$, which are trained to predict a denoised variant of their input $z_t$, which is a noisy version of $z$. The objective of instructional supervised training is:
\begin{equation}
\mathcal{L}=\mathbb{E}_{\mathcal{E}(\tilde{\boldsymbol{x}}), c, \epsilon \sim \mathcal{N}(0,I), t}\left[\| \epsilon-\epsilon_\theta\left(z_t, t, c\right) \|_2^2\right].
\end{equation}
HIVE proposes that optimizing a exponential reward weighted objective for fine-tuning diffusion models:
\begin{equation}
\mathcal{L}_{\mathrm{HIVE}}(\theta):=\mathbb{E}_{\mathcal{E}(\tilde{x}), c, \epsilon \sim \mathcal{N}(0,I), t}\left[\omega(\tilde{x}, c) \cdot\left\|\epsilon-\epsilon_\theta\left(z_t, t, c\right)\right\|_2^2\right],
\end{equation}
with $\omega(\tilde{x}, c)=\exp \left(r_\phi(\tilde{x}, c) / \beta\right)$ being the exponential reward weight for edited image $\tilde{x}$ and condition $c$, which is motivated by the closed form of optimal solution for RLHF in Eq. \eqref{Optimal Solution to RLHF}.

\section{Offline Alignment}
In this section, we present a detailed explanation for each offline preference tuning method, including SLiC-HF, DPO and its variants. In Table \ref{Summary on XPOs}, for simplicity, we include representative DPO variants and their final loss objectives. For each DPO variant, we conclude not only the resulting final objective or algorithm, but also both summarize the motivation or the direction the method contributed to for improvement over DPO.

\begin{table*}[!th]
\centering
\resizebox{0.9\textwidth}{!}{
\begin{tabular}{ll}
\toprule 
\textbf{Method} & \textbf{Objective}  \\
\midrule 
DPO & $-\log \sigma \left( \beta_\text{reg} \log \frac{\pi_\theta(y_w|x)}{\pi_{\text{ref}}(y_w|x)} - \beta_\text{reg} \log \frac{\pi_\theta(y_l|x)}{\pi_{\text{ref}}(y_l|x)}\right)$ \\ \midrule 
IPO & $ \left( \beta_\text{reg}\log \frac{\pi_\theta(y_w|x)}{\pi_{\text{ref}}(y_w|x)} - \beta_\text{reg}\log \frac{\pi_\theta(y_l|x)}{\pi_{\text{ref}}(y_l|x)} - \frac{1}{2} \right)^2 $ \\  
\midrule
$f$-DPO & $-\log \sigma\left(\beta_\text{reg} f^{\prime}\left(\frac{\pi_{\boldsymbol{\theta}}\left(y_w \mid x\right)}{\pi_{\mathrm{ref}}\left(y_w \mid x\right)}\right)-\beta_\text{reg} f^{\prime}\left(\frac{\pi_{\boldsymbol{\theta}}\left(y_l \mid x\right)}{\pi_{\mathrm{ref}}\left(y_l \mid x\right)}\right)\right)$  \\ 
\midrule
KTO & $-\lambda_w \sigma \left( \beta_\text{reg} \log \frac{\pi_\theta(y_w|x)}{\pi_{\text{ref}}(y_w|x)} - z_{\text{ref}} \right) -  \lambda_l \sigma \left( z_{\text{ref}} - \beta_\text{reg} \log \frac{\pi_\theta(y_l|x)}{\pi_{\text{ref}}(y_l|x)} \right),\,$ \\  
& $\text{where} \,\, z_{\text{ref}} = \mathbb{E}_{(x, y) \sim \mathcal{D}} \left[\beta_\text{reg} \text{KL}\left( \pi_\theta(y|x) || \pi_{\text{ref}}(y|x) \right)  \right]$ \\ 
\midrule
ODPO & $-\log \sigma \left( \beta_\text{reg} \log \frac{\pi_\theta(y_w|x)}{\pi_{\text{ref}}(y_w|x)} - \beta_\text{reg} \log \frac{\pi_\theta(y_l|x)}{\pi_{\text{ref}}(y_l|x)}-\Delta_r(x)\right)$\\  
\midrule
Mallows-DPO & $-\log \sigma  \left( \phi(x)\left[\beta_\text{reg} \log \frac{\pi_\theta(y_w|x)}{\pi_{\text{ref}}(y_w|x)} - \beta_\text{reg} \log \frac{\pi_\theta(y_l|x)}{\pi_{\text{ref}}(y_l|x)}\right] \right)$\\  
\midrule
R-DPO & $-\log \sigma \left( \beta_\text{reg} \log \frac{\pi_\theta(y_w|x)}{\pi_{\text{ref}}(y_w|x)} - \beta_\text{reg} \log \frac{\pi_\theta(y_l|x)}{\pi_{\text{ref}}(y_l|x)} - \left(\alpha |y_w| - \alpha |y_l| \right) \right)$ \\ \midrule
CPO & $-\log p_\theta(y_w|x)-\log \sigma \left( \beta_\text{reg} \log \pi_\theta(y_w|x)- \beta_\text{reg} \log \pi_\theta(y_l|x)\right)$  \\ 
\midrule
ORPO & $-\log p_\theta(y_w|x) - \lambda  \log \sigma \left(\log \frac{p_\theta(y_w|x)}{1 - p_\theta(y_w|x)} - \log \frac{p_\theta(y_l|x)}{1 - p_\theta(y_l|x)}  \right),\,$  \\  
& $\text{where} \,\, p_\theta(y|x) = \exp\left( \frac{1}{|y|} \log \pi_\theta(y|x) \right)$ \\  
\midrule
SimPO & $-\log \sigma  \left( \frac{\beta_\text{reg}}{|y_w|} \log \pi_\theta(y_w|x) - \frac{\beta_\text{reg}}{|y_l|} \log \pi_\theta(y_l|x) - \gamma \right)$ \\
\midrule
RainbowPO & $- \log\sigma\left(\phi(x)\left[\frac{\beta_\text{reg}}{\left|y^w\right|} \log \frac{\pi_\theta\left(y_w \mid x\right)}{\pi_\alpha\left(y_w \mid x\right)}-\frac{\beta_\text{reg}}{\left|y^l\right|} \log \frac{\pi_\theta\left(y_l \mid x\right)}{\pi_\alpha\left(y_l \mid x\right)}\right]\right)$ \\
\bottomrule
\end{tabular}
}
\label{Summary on XPOs}
\caption{Various preference optimization DPO objectives. The table is inspired from~\cite{meng2024simpo}.}
\end{table*}

\subsection{Offline Directed Preference Optimization (Offline DPO)}
\label{subsec:DPO}
One disadvantage of RLHF is that the RL step often requires substantial computational effort (e.g., to carry out the proximal policy optimization). DPO, recently proposed by \cite{rafailov2024direct}, suggested a possible way to bypass the reward modeling stage and avoid RL, and has attracted great attention. The key idea of DPO is the observation that given a reward function $r(x,y)$, the problem in Eq. \eqref{RLHF objective} has a closed-form solution:
\begin{equation}
\label{Optimal Solution to RLHF}
\pi_r(y \mid x)=\frac{1}{Z(x)} \pi_{\text {ref}}(y \mid x) \exp \left(\frac{1}{\beta_\text{reg}} r(x, y)\right),
\end{equation}
where $Z(x)=\sum_y \pi_{\text{ref}}(y \mid x) \exp \left(\frac{1}{\beta_\text{reg}} r(x, y)\right)$ is a normalizing constant.
Rearranging the terms, and plug in the ground truth reward $r^*$ with the optimal policy $\pi^*=\pi_{r^*}$ yield:
\begin{equation}
\label{reparameterization}
r^*(x, y)=\beta_\text{reg} \log \frac{\pi^*(y \mid x)}{\pi_{\text {ref}}(y \mid x)}+\beta_\text{reg} \log Z(x) .
\end{equation}
Through this change of variables, the latent reward $r^*(x,y)$ can be expressed in terms of the optimal
 policy $\pi^*(y \mid x)$,
the reference policy $\pi_{\text {ref }}(y \mid x)$
and a constant $Z^*(x)$.
Substituting this $r^*$ expression into Eq. \eqref{BT} yields:
\begin{equation}
\label{eq:ppi}
p^*\left(y_1 \succ y_2 \mid x\right)=\sigma\left(\beta_\text{reg} \log \frac{\pi^*\left(y_1 \mid x\right)}{\pi_{\text {ref }}\left(y_1 \mid x\right)}-\beta_\text{reg} \log \frac{\pi^*\left(y_2 \mid x\right)}{\pi_{\text {ref }}\left(y_2 \mid x\right)}\right),
\end{equation}
where $Z^*(x)$ cancels out and motivates the DPO objective:
\begin{multline}
\label{DPO objective}
\mathcal{L}_{\mathrm{DPO}}\left(\pi_{\theta} ; \pi_{\mathrm{ref}}\right)
:=\\
-\mathbb{E}_{\left(x, y_w, y_l\right) \sim \mathcal{D}}\left[\log \sigma\left(\beta_\text{reg} \log \frac{\pi_{\theta} \left(y_w \mid x\right)}{\pi_{\text {ref }}\left(y_w \mid x\right)}-\beta_\text{reg} \log \frac{\pi_{\theta} \left(y_l \mid x\right)}{\pi_{\text {ref }}\left(y_l \mid x\right)}\right)\right],
\end{multline}
which is a supervised learning problem, requiring much less computation than the RLHF. To understand the loss objective of DPO, we can further examine its gradient as following:
\begin{multline}
    \nabla_\theta \mathcal{L}_\text{DPO}(\pi_\theta;\pi_{ref}) = \\ -\beta_\text{reg}\mathbb{E}_{(x, y_w, y_l) \sim \mathcal{D}} \bigg[\underbrace{\sigma(\hat{r}_\theta(x, y_l) - \hat{r}_\theta (x, y_w))}_\text{higher weight when estimate is wrong}\bigg[\underbrace{\nabla_\theta\log \pi(y_w \mid x)}_\text{increase likelihood of $y_w$} \\
    - \underbrace{\nabla_\theta\log\pi(y_l \mid x)}_\text{decrease likelihood of $y_l$}\bigg]\bigg],
\end{multline}
in which
\begin{equation}
\hat{r}_\theta(x, y)=\beta_\text{reg} \log \frac{\pi_\theta(y \mid x)}{\pi_{\text {ref }}(y \mid x)},
\end{equation}
is called the implicit reward model for the policy $\pi_{\theta}$ in DPO.

\subsubsection{Identity Preference Optimization (IPO)}
For DPO variants, we first visit IPO, proposed in \cite{azar2024general}, motivated to bypass the assumption of Bredley-Terry model in the derivation of DPO (which comes from the reward modeling stage of RLHF). \cite{azar2024general} first propose a generic form of regularized optimization objective as:
\begin{equation}
\label{PsiPO objective}
\max _\pi \underset{\substack{x \sim \rho \\ y \sim \pi(. \mid x) \\ y^{\prime} \sim \mu(. \mid x)}}{\mathbb{E}}\left[\Psi\left(p^*\left(y \succ y^{\prime} \mid x\right)\right)\right]-\beta_\text{reg} D_{\mathrm{KL}}\left(\pi \| \pi_{\mathrm{ref}}\right)
\end{equation}
in which the new introduced function $\Psi$ is non-decreasing. They show that Eq. \eqref{PsiPO objective} shares the same optimality as DPO when taking $\Psi(q)=\log (q /(1-q))$ (notably this equivalence still needs  the Bradley-Terry model assumption). Furthermore, \cite{azar2024general} show that when $\Psi(x) = x$, i.e., when $\Psi$ is the identity mapping, Eq. \eqref{PsiPO objective} is equivalent to:
\begin{equation}
\label{IPO objective}
\mathcal{L}_{\mathrm{IPO}}\left(\pi_{\theta} ; \pi_{\mathrm{ref}}\right)
:=\mathbb{E}_{\left(x, y_w, y_l\right) \sim \mathcal{D}}\left(\beta_\text{reg} \log \frac{\pi_{\theta} \left(y_w \mid x\right)}{\pi_{\text {ref }}\left(y_w \mid x\right)}-\beta_\text{reg} \log \frac{\pi_{\theta} \left(y_l \mid x\right)}{\pi_{\text {ref }}\left(y_l \mid x\right)}-\frac{1}{2}\right)^2,
\end{equation}
if the offline dataset $\mathcal{D}$ is created by $x\sim \rho$ and $y,y^{\prime}\sim \mu$. Notice that the derivation of the objective in Eq. \eqref{IPO objective} does not acquire Bredley-Terry model, thus IPO is {\it preference model free}. In \cite{azar2024general}, it is also demonstrated through a synthetic bandit experiment that DPO can be prone to overfitting, while IPO could avoid this problem. In addition, also shows that online version of IPO \citep{calandriello2024human} (see details of online DPO in Section \ref{subsec:online dpo}) is indeed equivalent to Nash-MD proposed in Nash Learning from Human Feedback \citep{munos2023nash}.

\subsubsection{Rejection Sampling Optimization (RSO)}\label{subsubsection_rso}
RSO revisits the derivation of DPO and interpret the objective as a maximum likelihood estimator (MLE) of the optimal policy based on Eq. \eqref{eq:ppi}~\citep{liu2023statistical}. However, such a density estimation problem theoretically requires the datasets to be generated from the optimal policy instead of the SFT model in DPO. Thus, RSO algorithm is proposed to generate the datasets from the approximated optimal policy with an aid of a trained reward model $r_{\phi^*}$ and statistical rejection sampling, see in Algorithm \ref{Alg:RSO}.
\begin{table*}[!htbp]
\centering
\begin{minipage}{0.42\textwidth}
    \centering
    \begin{algorithm}[H]
    \begin{algorithmic}[1]
    \State \small Start with empty $\mathcal{Y} \gets \{\}$.
    \While{not enough samples in $\mathcal{Y}$}
        \State Generate $y \sim \pi_{\text{sft}}(y \mid x)$ that is not in $\mathcal{Y}$.
        \State Generate $u \sim U[0,1]$ and let \small{$M = \min \left\{m \mid m \geq \frac{\pi_{r_{\phi^*}}(y \mid x)}{\pi_{\text{sft}}(y \mid x)} \text{ for all } y \notin \mathcal{Y}\right\}$}.
        \If{$u < \frac{\pi_{r_{\phi^*}}(y \mid x)}{M \pi_{\text{sft}}(y \mid x)}$}
            \State Accept $y$ and add it to $\mathcal{Y}$.
        \Else 
        \State Reject $y$.
        \EndIf
    \EndWhile
    \end{algorithmic}
    \caption{RSO algorithm.}
    \label{Alg:RSO}
    \end{algorithm}
\end{minipage}
\hfill
\begin{minipage}{0.57\textwidth}
    \centering
    \includegraphics[width=.9\linewidth]{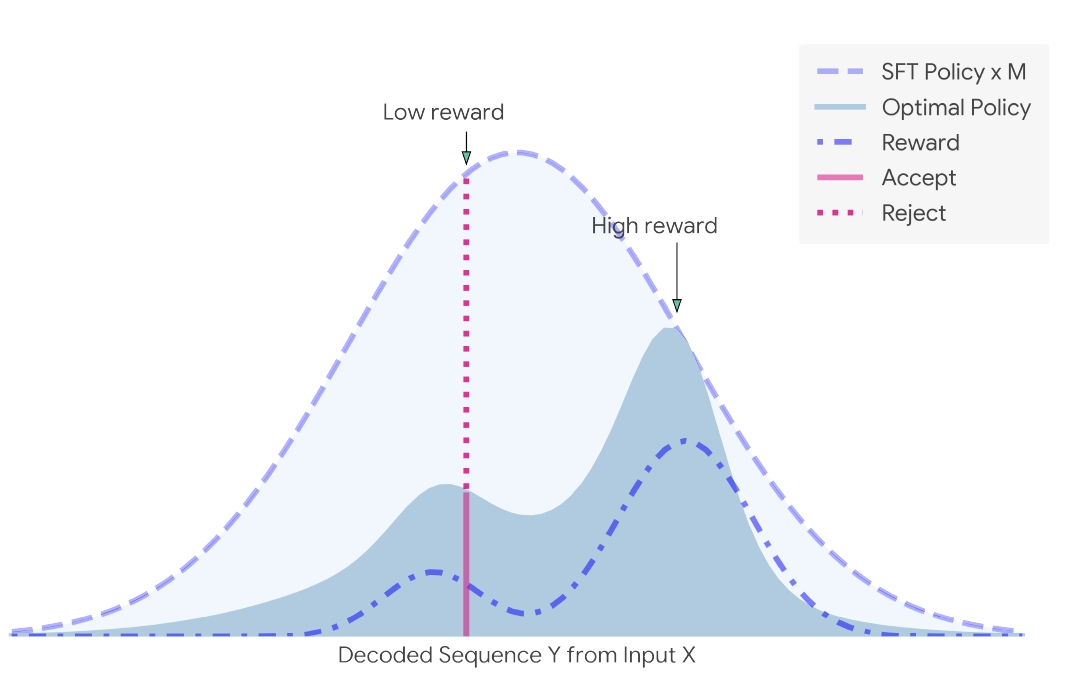}
    \caption{RSO illustration in \cite{liu2023statistical}.}
    \label{fig:RSO_demo}
\end{minipage}%
\end{table*}
Notice that $\pi_{r_{\phi^*}}$ is computed by Eq. \eqref{Optimal Solution to RLHF} with the learned reward model $r_{\phi^*}$.~\cite{liu2023statistical} show that this {\it distribution correction} could help improve the performance of DPO by utilizing the resampled preference dataset.
\begin{table*}[!ht]
\centering
\resizebox{0.95\textwidth}{!}{
\begin{minipage}{0.6\textwidth}
    \centering
    \begin{tabular}{llc}
    \toprule 
    \textbf{Method} & \textbf{Loss Function} & $f(x)$  \\
    \midrule 
    DPO & log logistic & $-\log\sigma(x)$\\ 
    \midrule 
    IPO & square & $(x-1)^2$\\  
    \midrule
    SLiC-HF & hinge loss & $\max(0,1-x)$\\ 
    \bottomrule
    \end{tabular}
    \caption{Unified perspective through loss function in \cite{liu2023statistical,tang2024generalized}.}
    \label{table:GPO-unified-perspective}
\end{minipage}
\begin{minipage}{0.4\textwidth}
    \centering
    \includegraphics[width=0.7\linewidth]{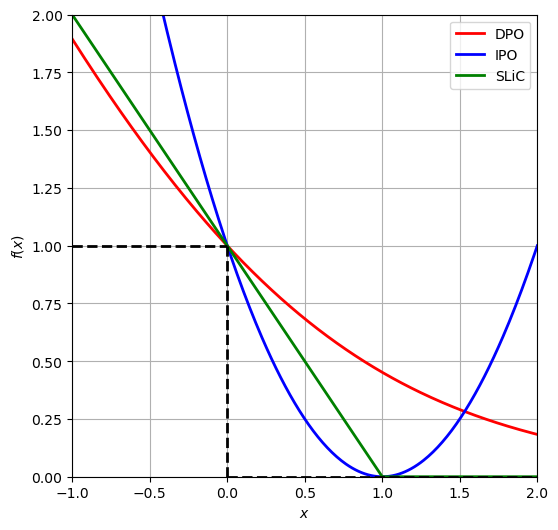}
    \caption{Loss function comparison in \cite{tang2024generalized}.}
    \label{fig:loss-comparison}
\end{minipage}
}
\end{table*}

In addition, RSO also unifies DPO and (normalized) SLiC-HF from the perspective of {\it loss function}; similar unified perspective also appeared in GPO \citep{tang2024generalized} (see e.g., in Table 1 of it):
\begin{equation}
\label{GPO objective}
\mathcal{L}_{\mathrm{GPO}}\left(\pi_{\theta} ; \pi_{\mathrm{ref}}\right)
:=\mathbb{E}_{\left(x, y_w, y_l\right) \sim \mathcal{D}}\left[f\left(\beta_\text{reg} \log \frac{\pi_{\theta} \left(y_w \mid x\right)}{\pi_{\text {ref }}\left(y_w \mid x\right)}-\beta_\text{reg} \log \frac{\pi_{\theta} \left(y_l \mid x\right)}{\pi_{\text {ref }}\left(y_l \mid x\right)}\right)\right],
\end{equation}
for any convex function $f$, like in Table \ref{table:GPO-unified-perspective}. GPO further provides an analysis of this formulation from an policy improvement and policy regularization trade-off. Applying Taylor Expansion of the form above yields: 
\begin{equation}
\mathcal{L}_{\mathrm{GPO}}\left(\pi_{\theta} ; \pi_{\mathrm{ref}}\right)=f(0)+\beta_\text{reg} \underbrace{f^{\prime}(0)}_{<0}\underbrace{\mathbb{E}_{\left(x, y_w, y_l\right) \sim \mathcal{D}}\left[\rho_{\theta}\right]}_\text{ optimization}+\frac{1}{2}\beta_\text{reg}^2 \underbrace{f^{\prime\prime}(0)}_{>0}\underbrace{\mathbb{E}_{\left(x, y_w, y_l\right) \sim \mathcal{D}}\left[\rho^2_{\theta}\right]}_\text{ regularization},
\end{equation}
in which $\rho_{\theta}=\log \frac{\pi_{\theta} \left(y_w \mid x\right)}{\pi_{\text {ref }}\left(y_w \mid x\right)}-\log \frac{\pi_{\theta} \left(y_l \mid x\right)}{\pi_{\text {ref }}\left(y_l \mid x\right)}$ denotes `implicit reward difference'. 

\subsubsection{\texorpdfstring{$f$-DPO}{f-DPO}}
DPO is derived from the RLHF objective which utilized the (reverse) KL divergence to prevent the deviation of new models from old models. $f$-DPO in~\cite{wang2023beyond} consider extending this statistical distance to general $f$-divergence. Concretely, for two probability distribution $P$ and $Q$ with probability density function $p$ and $q$ respectively, $f$-divergence is defined as:
\begin{equation}
D_f(P||Q)=\mathbb{E}_{q(x)}\left[f\left(\frac{p(x)}{q(x)}\right)\right],
\end{equation}
and reverse KL divergence is a special case when taking $f(x)=x\log(x)$.
\cite{wang2023beyond} first show that through a first order condition of optimality / KKT and the similar change of variable technique in DPO, the RLHF objective with a $f$-divergence
\begin{align}
    \label{RLHF objective with f-divergence}
    \mathcal{L}_{\text{RLHF}-f}(\phi) = \mathbb{E}_{x\sim \mathcal{D}, y\sim{\pi(\cdot\mid x)}}\left[r_{\phi^*}(x,y)-\beta_\text{reg} D_f(\pi(\cdot\mid x)\mid \pi_{\text{ref}}(\cdot\mid x))\right],
\end{align}
could yield the $f$-DPO objective:
\begin{multline}
\label{f-DPO objective}
\mathcal{L}_{f-\mathrm{DPO}}\left(\pi_{\theta} ; \pi_{\mathrm{ref}}\right)=\\
\mathbb{E}_{\left(x, y_w, y_l\right) \sim \mathcal{D}}\left[-\log \sigma\left(\beta_\text{reg} f^{\prime}\left(\frac{\pi_{\theta}\left(y_w \mid x\right)}{\pi_{\mathrm{ref}}\left(y_w \mid x\right)}\right)-\beta_\text{reg} f^{\prime}\left(\frac{\pi_{\theta}\left(y_l \mid x\right)}{\pi_{\mathrm{ref}}\left(y_l \mid x\right)}\right)\right)\right].
\end{multline}
Special cases of Eq. \eqref{f-DPO objective} are when taking $f$ divergence as $\alpha$-divergence and JS-divergence, and \cite{wang2023beyond} further argue that JS-divergence could possibly yield a better diversity and accuracy tradeoff than reverse KL, through small-scale experiments on e.g., IMDB controllable generation and fine-tuning Pythia 2.8B on Anthropic HH dataset.

\subsubsection{Kahneman-Tversky Optimization (KTO)}
KTO~\citep{ethayarajh2024kto} is motivated to address the need of pairwise preferences datasets in DPO, which can be scarce and expensive. Instead of maximizing the log-likelihood of preferences in DPO and inspired by Kahneman \& Tversky’s prospect theory, KTO proposes to minimize a human-aware loss function (HALO) that represents the utility of generations and also takes into account the human nature of loss aversion. The resulting KTO objective decouples the pair-preferences into two separate terms that are further linearly combined:
\begin{multline}
\label{KTO objective}
\mathcal{L}_{\mathrm{KTO}}\left(\pi_{\theta} ; \pi_{\mathrm{ref}}\right)
:=\\
-\mathbb{E}_{\left(x, y_w, y_l\right) \sim \mathcal{D}}\left[\lambda_w \sigma ( \beta_\text{reg} \log \frac{\pi_\theta(y_w|x)}{\pi_{\text{ref}}(y_w|x)} - z_{\text{ref}} ) +  \lambda_l \sigma ( z_{\text{ref}} - \beta_\text{reg} \log \frac{\pi_\theta(y_l|x)}{\pi_{\text{ref}}(y_l|x)} )\right],
\end{multline}
where $z_{\text{ref}} = \mathbb{E}_{(x, y) \sim \mathcal{D}} \left[\beta_\text{reg} \text{KL}\left( \pi_\theta(y|x) || \pi_{\text{ref}}(y|x) \right)  \right]$ acts like a subjective value and $\lambda_w,\lambda_l$ are additonal hyper-parameters to be tuned. If there is only desired/undesired answer, the KTO objective will thus have only one term, which makes it {\it pairwise preference data free}.

\subsubsection{Offset DPO (ODPO)}
DPO objective cannot reflect the {\it significance} of the preference pairs i.e., the extent $y_w$ is preferred to $y_l$, and ODPO~\citep{amini2024direct} propose to add a margin to capture this significance; they model this margin, or they call offset $\Delta_r$ as a monotonically increasing function $f(\cdot)$ of the difference between the scores associated with the responses:
\begin{equation}
\Delta_r(x,y_w,y_l)=\alpha f\left(\operatorname{score}\left(x, y_w\right)-\operatorname{score}\left(x, y_l\right)\right),
\end{equation}
where $\alpha$ is a hyper-parameter that controls the extent to which an offset should be enforced. The resulting objective becomes:
\begin{multline}
\label{ODPO objective}
\mathcal{L}_{\mathrm{ODPO}}\left(\pi_{\theta} ; \pi_{\mathrm{ref}}\right)
:=\\
-\mathbb{E}_{\left(x, y_w, y_l\right) \sim \mathcal{D}}\left[\log \sigma \left( \beta_\text{reg} \log \frac{\pi_\theta(y_w|x)}{\pi_{\text{ref}}(y_w|x)} - \beta_\text{reg} \log \frac{\pi_\theta(y_l|x)}{\pi_{\text{ref}}(y_l|x)}-\Delta_r(x,y_w,y_l)\right)\right].
\end{multline}

\subsubsection{Mallows-DPO}
Mallows-DPO \citep{chen2024mallows} is motivated by DPO's lack of capability to characterize the diversity of human preferences. Inspired by Mallows Ranking Model (opposed to Bredley-Terry in RLHF and DPO) which has a natural carrier of a dispersion index, Mallows-DPO pays attention to the {\it dispersion} of the preferences: when human tends to agree about the answer to a certain question, e.g., `$1+1=?$', the preference dispersion will be small; however, the dispersion will be large for answer to a general open question. \cite{chen2024mallows} propose a contextual scaled objective derived from MLE under Mallows: compared to DPO that puts equal weights on each prompt and preference pairs, the resulting Mallows-DPO adds a contextual scaling factor $\phi(x)$ that represents this dispersion of the preferences of answers to each prompt $x$:
\begin{multline}
\label{Mallows-DPO objective}
\mathcal{L}_{\mathrm{Mallows-DPO}}\left(\pi_{\theta} ; \pi_{\mathrm{ref}}\right)
:=\\
-\mathbb{E}_{\left(x, y_w, y_l\right) \sim \mathcal{D}}\left[\log \sigma  \left( \phi(x)\left[\beta_\text{reg} \log \frac{\pi_\theta(y_w|x)}{\pi_{\text{ref}}(y_w|x)} - \beta_\text{reg} \log \frac{\pi_\theta(y_l|x)}{\pi_{\text{ref}}(y_l|x)}\right] \right)\right].
\end{multline}
To compute this dispersion, Mallows-DPO provided a direct approach by using a normalized predictive entropy of preference pairs $\{y_i^w,y_i^l\}_{i=1, \ldots, N}$ with $N=\max(|y^w|,|y^l|)$:
\begin{equation}
\label{eqn:mallows dpo dispersion estimator}
\phi(x)=-\log\left(\frac{\frac{1}{2} \sum_{i=1}^{N-1}\left[H_{\pi_{\text{ref}}}(Y_{i+1}\mid Y_i=y^w_i)+H_{\pi_{\text{ref}}}(Y_{i+1}\mid Y_i=y^l_i)\right]}{\log(n)}\right).
\end{equation}
To illustrate the effect of this additional term, when dispersion is high: $\phi(x)$ in Eq. \eqref{eqn:mallows dpo dispersion estimator} will be close to 0, and Mallows-DPO will put less weights on the corresponding preference pairs in the optimization objective to prevent from overfitting; In contrast, when dispersion is low, Mallows-DPO put more weights in the preference optimization objective, for which $\phi(x)$ is large and will lead to stronger effect of alignment. 

\subsubsection{LR-DPO}
LR-DPO~\citep{park2024disentangling}, DPO with length regularization is motivated to address the problem of verbosity in the DPO setting. LR-DPO proposed a simple regularization strategy that prevents
length exploitation by penalizing the rewards with length of the generation in standard RLHF objective:
\begin{align}
    \label{RLHF objective length regularized}
    \mathcal{L}_{\text{LR-RLHF}}(\phi) = \mathbb{E}_{x\sim \mathcal{D}, y\sim{\pi(\cdot\mid x)}}\left[r_{\phi^*}(x,y)-\alpha |y|-\beta_\text{reg} \operatorname{KL}(\pi(\cdot\mid x)\mid \pi_{\text{ref}}(\cdot\mid x))\right],
\end{align}
in which $\alpha$ is a hyper-parameter that controls the extent of length regularization. Eq. \eqref{RLHF objective length regularized} thus similarly yields a supervised learning objective referred as DPO with length regularization:
\begin{multline}
\label{R-DPO objective}
\mathcal{L}_{\mathrm{LR-DPO}}\left(\pi_{\theta} ; \pi_{\mathrm{ref}}\right) \\
=-\underset{\left(x, y_w, y_l\right) \sim \mathcal{D}}{\mathbb{E}}\log \sigma \left( \beta_\text{reg} \log \frac{\pi_\theta(y_w|x)}{\pi_{\text{ref}}(y_w|x)} - \beta_\text{reg} \log \frac{\pi_\theta(y_l|x)}{\pi_{\text{ref}}(y_l|x)} -(\alpha|y_w|-\alpha|y_l|)\right).
\end{multline}
\cite{park2024disentangling} further show that this can effectively improve model quality by addressing the verbosity issue.

\subsubsection{Contrastive Preference Optimization (CPO)}
CPO~\citep{xu2024contrastive} is motivated to improve the memory and speed efficiency of DPO by neglecting the reference policy, further accompanied by a SFT loss term:
\begin{equation}
\label{CPO objective}
\mathcal{L}_{\mathrm{CPO}}\left(\pi_{\theta}\right)
:=-\mathbb{E}_{\left(x, y_w, y_l\right) \sim \mathcal{D}}\left[\log p_\theta(y_w|x)+\log \sigma \left( \beta_\text{reg} \log \frac{\pi_\theta(y_w|x)}{\pi_\theta(y_l|x)} \right)\right].
\end{equation}

\subsubsection{Odds Ratio Preference Optimization (ORPO)}
Opposed to maximizing the likelihood ratios of winning and losing answers in the preference pair in DPO, ORPO \citep{hong2024orpo} propose that odds ratio can be a more sensible choice. 
\begin{multline}
\label{ORPO objective}
\mathcal{L}_{\mathrm{ORPO}}\left(\pi_{\theta}\right)
:=\\
-\mathbb{E}_{\left(x, y_w, y_l\right) \sim \mathcal{D}}\left[\log p_\theta(y_w|x) + \lambda  \log \sigma \left(\log \frac{p_\theta(y_w|x)}{1 - p_\theta(y_w|x)} - \log \frac{p_\theta(y_l|x)}{1 - p_\theta(y_l|x)}  \right)\right].
\end{multline} 
where $p_\theta(y|x) = \exp\left( \frac{1}{|y|} \log \pi_\theta(y|x) \right)$. ORPO is similar to CPO in the sense that it is also reference model free and combined with a SFT loss; in addition, notably that ORPO also adopts a form of length regularization by normalizing the likelihoods with respect to the length, as in the definition of $p_\theta(y|x)$; finally, they compute odds ratio instead of the original likelihood ratio.

\subsubsection{SimPO}
SimPO \citep{meng2024simpo} propose a simple yet effective objective that is claimed to match or even outperform the performance of DPO: 
\begin{equation}
\label{SimPO objective}
\mathcal{L}_{\mathrm{SimPO}}\left(\pi_{\theta}\right)
:=-\mathbb{E}_{\left(x, y_w, y_l\right) \sim \mathcal{D}}\left[\log \sigma  \left( \frac{\beta_\text{reg}}{|y_w|} \log \pi_\theta(y_w|x) - \frac{\beta_\text{reg}}{|y_l|} \log \pi_\theta(y_l|x) - \gamma \right)\right],
\end{equation} 
where $\gamma$ is introduced as a target reward margin to help separating the winning and losing responses. SimPO is similar to CPO in the sense of being reference model free; it also adopted the length normalization for the likelihoods as in ORPO; finally, it additionally introduced a constant margin to be tuned that could help to further improve the performance by encouraging a larger difference between the normalized likelihoods.

\subsubsection{RainbowPO}
Inspired by the paper Rainbow on improving DQN for better performance, RainbowPO \citep{zhao2024rainbowpo} demystifies the effectiveness of existing DPO variants by categorizing their key components into several broad directions, and integrate the identified effective components into a single cohesive objective: 
\begin{align}
\mathcal{L}_{\text{RainbowPO}}\left(\pi_\theta ; \pi_{\text {ref }}\right)=-\underset{\left(x, y_w, y_l\right) \sim \mathcal{D}}{\mathbb{E}} f\left[\phi(x)\left(\frac{\beta}{\left|y^w\right|^\eta} \log \frac{\pi_\theta\left(y_w \mid x\right)}{\pi_\alpha\left(y_w \mid x\right)}-\frac{\beta}{\left|y^l\right|^\eta} \log \frac{\pi_\theta\left(y_l \mid x\right)}{\pi_\alpha\left(y_l \mid x\right)}\right)\right],
\end{align}
in which $\eta \in\{0,1\}$, and $\pi_\alpha$ is referred to a mixing policy mechanism they propose for formulating a better reference policy by mixing policy $\pi_{\text{ref}}$ and $\pi_{\gamma}$, defined as:
\begin{equation}
\pi_\alpha(y \mid x) \propto \pi_{\mathrm{ref}}^\alpha(y \mid x) \cdot \pi_\gamma^{1-\alpha}(y \mid x),
\end{equation}
and $\pi_\gamma$ is a policy which assumes to exist (which can be understood as the reference policy taken by SimPO \citep{meng2024simpo}), such that the model is perfect at distinguishing the preference pairs in the dataset:
\begin{equation}
\pi_\gamma\left(y_w \mid x\right)^{1 /\left|y^\omega\right|}/ \pi_\gamma\left(y_l \mid x\right)^{1 /\left|y^l\right|}=\exp (\gamma),
\end{equation}
for any prompt $x$. \cite{zhao2024rainbowpo} show that optimizing such generic objective can yield the best performance on downstream task of tuning Llama3-8B-Instruct for instruction-following capabilities, benefiting from composition of effective elements.

\subsection{Multi-Modal Models}
\subsubsection{Diffusion-DPO}
Diffusion-DPO~\citep{wallace2024diffusion} is adapting DPO to diffusion models. It uses a fixed dataset and each example contains a prompt and a pairs of images generated from a reference model with human label. Similar to RL for diffusion, the goal is still to align the base diffusion models to human preferences. The derivation is similar to RL framework for diffusion in DDPO and DPOK, and also DPO for Language Models: 
\begin{equation}
\mathcal{L}(\theta)=-\mathbb{E}_{\left(x_0^w, x_0^l\right) \sim \mathcal{D}} \log \sigma(\beta_\text{reg} \mathbb{E}_{\substack{x_{1: T}^w \sim p_\theta\left(x_{1: T}^w \mid x_0^w\right) \\
x_{1: T} \sim p_\theta\left(x_{1: T}^l \mid x_0^l\right)}}\left[\log \frac{p_\theta\left(x_{0: T}^w\right)}{p_{\text {ref }}\left(x_{0: T}^w\right)}-\log \frac{p_\theta\left(x_{0: T}^l\right)}{p_{\text {ref }}\left(x_{0: T}^l\right)}\right]).
\end{equation}
However, the main concern left is that the likelihood term of the generations $p_\theta\left(x_{0: T}\right)$ is not tractable if only given generation $x_0$. \cite{wallace2024diffusion} further propose to use the forward process $q\left(x_{1: T} \mid x_0\right)$ of diffusion to match the distribution of backward process $p_\theta\left(x_{1: T} \mid x_0\right)$, and yield the final DPO-Diffusion objective:
\begin{multline}
    L_{\text{DPO-diffusion}}(\theta)=-\mathbb{E}_{\left(x_0^w, x_0^l\right) \sim \mathcal{D}, t \sim \mathcal{U}\left[0, T\right], x_t^w \sim q\left(x_t^w \mid x_0^w\right), x_t^l \sim q\left(x_t^l \mid x_0^l\right)} \log \sigma(-\beta_\text{reg} T\\
    \left[\mathrm{KL}\left(q\left(x_{t-1}^w \mid x_{0, t}^w\right) \| p_\theta\left(x_{t-1}^w \mid x_t^w\right)\right) -\mathrm{KL}\left(q\left(x_{t-1}^w \mid x_{0, t}^w\right)\| p_{\text{ref}}\left(x_{t-1}^w \mid x_t^w\right)\right)\right. \\
    \left. -\mathrm{KL}\left(q(x_{t-1}^l \mid x_{0, t}^l) \| p_\theta(x_{t-1}^l \mid x_t^l)\right) +\mathrm{KL}\left(q(x_{t-1}^l \mid x_{0, t}^l) \| p_{\text{ref}}(x_{t-1}^l \mid x_t^l)\right)\right]),
\end{multline}
with each term can be readily computed.

\subsubsection{POVID} 
POVID~\citep{zhou2024aligning} proposes a method for performing preference optimization in visual language models (VLLM) with synthetically generated preferences. This is mainly aimed at attenuating the hallucination problems in VLLMs that arises due to lack of alignment between the language and vision modalities. Specifically, the authors use the ground-truth instructions as the preferred response and employ a two-stage
approach to generate dis-preferred responses: first, use GPT-4V to inject hallucinatory texts
into the preferred responses, and second, add diffusion noise to the
image to trigger the inherent hallucination behavior of the VLLM by making the image difficult for the VLLM to understand. Both the strategies are merged together in an reformulation of the DPO loss as: 
\begin{equation}
\begin{aligned}
\label{Povid objective}
\mathcal{L}_\text{POVID}(\pi_{\theta};\pi_{ref})
&= - \mathbb{E}_{(x, y_w, y_l) \sim \mathcal{D}} \left[ \log \sigma \left(  \alpha \log \frac{\pi_\theta (y_w \mid x)}{ \pi_{ref}(y_w \mid x)} \right. \right. \\
& ~~~ - \left. \left. \left( \beta_{\text{reg}_1} \log \frac{\pi_\theta (y^t_l \mid x)}{ \pi_{ref}(y^t_l \mid x)} + \beta_{\text{reg}_2} \log \frac{\pi_\theta (y_l^n \mid x^n)}{ \pi_{ref}(y_l^n \mid x^n)} \right) \right)\right],
\end{aligned}
\end{equation} 
where $\alpha, \beta_{\text{reg}_1}, \beta_{\text{reg}_2}$ are coefficients for balancing preferred responses ($y_w)$ and dispreferred responses ($y^t_l, y^n_l$). $y^t_l$ indicates the dispreferred response generated using GPT-4V, and $y^n_l$ denotes the dispreferred response generated using the noisy image $x^n$. 

\subsection{Sequence Likelihood Calibration (SLiC-HF)}
SLiC-HF~\citep{zhao2023slic} uses a sequence level contrastive learning training method to align the model's sequence likelihood over the decoded sequences by measuring their similarity with given reference sequences. The main reason to use a contrastive objective is to put more loss on negative sequence compared to positive sequences such that model puts more probability mass on generating positive sequences. Further, this specific formulation allows the use of human preference for ranking directly by using offline policy preference data $\mathcal{D}$ or by training a separate predictive ranking model on offline data. SLiC-HF obtains a supervised fine-tuned model $\pi_{\theta_{\text{ref}}} (y\mid x)$, which we denote as the reference model for consistency with RLHF pipelines on a reference dataset $(x,y_{\text{target}}) \sim \mathcal{D}$. The preference datasets $\{ y_w,y_l\}_m$ is formulated by uniformly drawing answer pairs from $\pi_{\theta_{\text{ref}}} (\cdot \mid x)$ and ranking them by their similarity (from a score computed by a pre-trained model denoted as $s(y,y_{\text{ref}};x)$) to the target answer $y_{\text{ref}}$. The step after is to align the SFT model's sequence likelihood using the SLiC loss~\citep{zhao2022calibrating}: 
\begin{equation}
\label{SLiC loss}
\mathcal{L}_{\text{SLiC}}{(\pi_{\theta} ; \pi_{\mathrm{ref}})}=\sum L^{\text{cal}} \left(\theta, x,y_{\text{target}}, \{ y_w,y_l\}_m \right) + \lambda L^{\text{reg}}\left(\theta, \theta_{\text{ref}}; x,y_{\text{target}}\right),
\end{equation}
in which $L^{\text{cal}}$ is the calibration loss from SLiC and $L^{\text{reg}}$ is the regularization loss to prevent the aligned model stray away from the SFT model. Taking a special case of $L^{\text{cal}}$ and $L^{\text{reg}}$ to be a rank calibration loss and cross entropy loss respectively, Eq. \eqref{SLiC loss} becomes: 
\begin{equation}
\label{SLiC HF loss}
\mathcal{L}_{\text{SLiC}}{(\pi_{\theta} ; \pi_{\mathrm{ref}})}= \underbrace{\max \left( 0, \delta - \log \pi_{\theta} (y_w| x) + \log \pi_{\theta} (y_l|x) \right)}_{\text{rank calibration loss}} \underbrace{-\lambda \log \pi_\theta ({y_{\text{ref}}| x})}_{regularization},
\end{equation}
where, in the first term of calibration loss, we are maximizing the likelihood corresponding to the positive sequence $y_w$ and minimizing negative sequence $y_l$ and the margin $\delta$ is a hyper-parameter represents which can be a constant or prompt dependent score/rank difference; the second term is just standard SFT loss. As a remark, one can use a secondary reward model, opposed to the similarity function in SLiC, trained on human preference data to classify positive or negative pairs $(y_w,y_l)$. 

\section{Combined Policies and Sampling-Agnostic Alignment}
In this section, we explore some other directions proposed in literature for improving the effectiveness of human preference tuning. We discuss ExPO~\citep{zheng2024weak}, which proposed that combining two aligned models by extrapolating from their weights could enhance the alignment quality of the model; we discuss P3O~\citep{fakoor2020p3o}, which utilized both on-policy and off-policy sampling; we also introduce sampling-agnostic alignment methods that can be applied to both off-policy and on-policy approaches.

\subsection{ExPO}
ExPO~\citep{zheng2024weak} provides a simple and training-free method for enhancing the alignment of large language models (LLMs) with human preferences. The core insight behind ExPO is that a model trained with DPO/RLHF can be viewed as an interpolation between two models with differing strengths. By leveraging this concept, one can potentially extrapolate a stronger model if the other two models are available. Specifically, if we denote the model $\pi_\text{ExPO}$ as the interpolation of two other models, $\pi_a$ and $\pi_b$, which may be trained using different alignment methods and datasets, ExPO assumes that combining these models will yield improved alignment. The stronger, better-aligned model $\pi_{ExPO}$ is then obtained by extrapolating from the weights of these two relatively weaker models (which is reminiscent to Model Soups \citep{wortsman2022model}), as formulated below:
\begin{align}
\pi_\text{ExPO}=(1+\alpha) \pi_{a}-\alpha \pi_{b}=\pi_{a}+\alpha\left(\pi_{a}-\pi_{b}\right)=\pi_{a}+\alpha \Delta \pi.
\end{align}
This method is shown to work when $\pi_a$ and $\pi_b$ are a stronger model from a combination of SFT model and a model further preference trained on top of it respectively. However in naive cases of choosing arbitrary $\pi_a$ and $\pi_b$, it has shown to cause model collapse or degradation. Nevertheless broader applicability of this approach requires further research.

\subsection{Policy-on Policy-off Policy Optimization (P3O)}
P3O~\citep{fakoor2020p3o} is a simple and effective algorithm that uses the effective sample size to automatically manage the combination of on-policy and off-policy optimization. It performs gradient ascent using the gradient. \cite{fakoor2020p3o} describe how P3O integrates the following on-policy update with the off-policy update:
\begin{align}
\nabla_\theta^{\mathrm{on}} J\left(\pi_\theta\right)&=\underset{s \sim d^\pi \theta, a \sim \pi_\theta}{\mathbb{E}}\left[g\left(\pi_\theta\right)\right], \\
\nabla_\theta^{\mathrm{off}} J\left(\pi_\theta\right)&=\underset{s \sim d^\beta_\text{reg}, a \sim \beta_\text{reg}}{\mathbb{E}}\left[\bar{\rho}_c g\left(\pi_\theta\right)\right],
\end{align}
where $\pi_\theta$ denotes a policy that is parameterized by parameters $\theta \in \mathbb{R}^n$, and $q^{\pi_\theta}$ and $v^{\pi_\theta}$ denote a parameterization of the state-action and state-only value functions, respectively. It is also denoted the baselined policy gradient integrand in short by following:
\begin{align}
g\left(\pi_\theta\right) &= \hat{A}^{\pi_\theta}(s, a) \nabla_\theta \log \pi_\theta(a \mid s), \\
\hat{A}^{\pi_\theta}(s, a)&=\hat{q}^{\pi_\theta}(s, a)-\hat{v}^{\pi_\theta}(s).
\end{align}
It forms a unified policy optimization as following:
\begin{align}
\underset{s \sim d^{\pi_\theta}, a \sim \pi_\theta}{\mathbb{E}} & {\left[g\left(\pi_\theta\right)\right]+\underset{s \sim d^\beta_\text{reg}, a \sim \beta_\text{reg}}{\mathbb{E}}\left[\bar{\rho}_c g\left(\pi_\theta\right)\right] } -\lambda \nabla_\theta \underset{s \sim d^\beta_\text{reg}, a \sim \beta_\text{reg}}{\mathbb{E}} \operatorname{KL}\left(\beta_\text{reg}(\cdot \mid s) \| \pi_\theta(\cdot \mid s)\right).
\end{align}
The first term above is the standard on-policy gradient. The second term is the off-policy policy gradient with truncation of the IS ratio using a constant $c$ while the third term allows explicit control of the deviation of the target policy $\pi_\theta$ from $\beta_\text{reg}$. 
Further, the KL-divergence term can be rewritten as $\underset{s \sim d^\pi \theta, a \sim \pi_\theta}{\mathbb{E}}[\log \rho]$ and therefore minimizes the importance ratio $\rho$ over the entire replay buffer $\beta_\text{reg}$. There are two hyper-parameters in the P3O gradient: the IS ratio threshold $c$ and the KL regularization co-efficient $\lambda$. 

\subsection{Reinforced Token Optimization (RTO)}

Standard RLHF and DPO's reward models are all based on the whole generation, thus the whole pipeline is in some sense closer to bandit instead of classical MDP based RL. Inspired by that nature of auto-regressive models is next token prediction , RTO~\citep{zhong2024dpo} derives a {\it token-wise} reward function from preference data and conducts policy optimization using this learned reward signal. Broadly, RTO formulates the optimization problem as an MDP and involves two primary steps: (1) learning a token-wise reward from preference data, and (2) optimizing this reward through RL training methods like PPO. 

\paragraph{Theoretical Version.}
Consider the offline setting by assuming that we have an offline dataset $\mathcal{D} = \{(\tau^w, \tau^l)\}$ that contains several trajectory pairs, where 
$\tau^w = \{(s_h^w, a_h^w)\}_{h=1}^H$ is preferred over $\tau^l = \{(s_h^l, a_h^l)\}_{h=1}^H$. Each pair of trajectories shares the same initial state (i.e., $s_1^w = s_1^l$), but differs in the subsequent tokens. RTO computes the maximum likelihood estimator $\theta_{mle}$ based on $\mathcal{D}$ by maximizing the log likelihood and calculates the pessimistic reward $\hat{r}$ via token-wise reward learning. The output of the algorithm is policy $\hat{\pi}$.

\paragraph{Practical Version}
Similar to learning the reward model in RLHF, the key challenge left is to learn the token-wise reward from the offline data. For sentence level reward, popular frameworks outlined in InstructGPT \citep{ouyang2022training}, Claude \citep{bai2022training}, and LLaMA2 \citep{touvron2023llama} replace the last layer of the LLM with a linear layer for a scalar output and maximize the log-likelihood, which thus cannot be naively used for token-level reward. RTO observes that, given a trajectory $\tau = \{(s_h, a_h)\}_{h=1}^H$, denoting $\pi_{\beta_\text{reg}}^{*}(a|s) = \exp\{ (Q_{\beta_{\text{reg}}}^*(s, a) - V_{\beta_\text{reg}}^*(s))/\beta_\text{reg} \}$ as the optimal policy, the KL regularization can be rewritten as:
\begin{align}
\sum_{h=1}^H \beta_\text{reg} \log \frac{\pi_{\beta_\text{reg}}^*(a_h|s_h)}{\pi_{\mathrm{ref}}(a_h|s_h)} &= \sum_{h = 1}^H \big(Q_{\beta_\text{reg}}^*(s_h, a_h) - V_{\beta_\text{reg}}^*(s_h) - \log \pi_{\mathrm{ref}}(a_h|s_h) \big) \notag \\
& = \sum_{h = 1}^H r(s_h, a_h) - V_{\beta_\text{reg}}^*(s_1) \\
& + \underbrace{\sum_{h = 1}^{H-1} \big( \mathbb{E}_{s' \sim \mathcal{P}(\cdot|s_h, a_h)}[V_{\beta_\text{reg}}^*(s')] - V_{\beta_\text{reg}}^*(s_{h+1}) \big)}_{(\star)},
\end{align}
in which the second equality follows from the fact that:
\begin{align}
Q_{\beta_\text{reg}}^{\pi}(s, a) = r_{\beta_\text{reg}}(s, a) + \mathbb{E}_{s' \sim \mathcal{P}(\cdot|s, a)}[V_{\beta_{\text{reg}^\pi}}(s')],
\end{align}
with $r_{\beta_{\text{reg}}}(s, a) = r(s, a) + \beta_\text{reg} \log \pi_{\mathrm{ref}}(a|s)$. RTO focuses on the typical LLM generation scenario where the transition kernel is deterministic. Then, $(\star) = 0$, yielding that 
\begin{align}
\sum_{h = 1}^H r(s_h, a_h) = \sum_{h=1}^H \beta_\text{reg} \log \frac{\pi_{\beta_\text{reg}}^*(a_h|s_h)}{\pi_{\mathrm{ref}}(a_h|s_h)} + V_{\beta_\text{reg}}^*(s_1) .
\end{align}
Building upon this result and combining it with the definition of the BT model, for any trajectory pair $\{\tau^j =  \{(s_{h}^j, a_{h}^j)\}_{h=1}^H\}_{j=1}^2$ satisfying $s_1^1 = s_1^2$, we have:
\begin{equation}
\begin{aligned}
    \mathbb{P}(\tau^1 \succ \tau^2) &= \sigma\left( \sum_{h=1}^H r(s_h^1, a_h^1) - \sum_{h=1}^H r(s_h^2, a_h^2)\right) \\
    &= \sigma\left( \sum_{h=1}^H \beta_\text{reg} \log \frac{\pi_{\beta_\text{reg}}^*(a_h^1|s_h^1)}{\pi_{\mathrm{ref}}(a_h^1|s_h^1)} - \sum_{h=1}^H \beta_\text{reg} \log \frac{\pi_{\beta_\text{reg}}^*(a_h^2|s_h^2)}{\pi_{\mathrm{ref}}(a_h^2|s_h^2)}\right).
\end{aligned}
\end{equation}
Similar to the bandit setting where the learning objective is equivalent to a BT model with sentence-wise reward $r^*(x, y) = \beta_\text{reg} \log \frac{\pi_{\beta_\text{reg}}^*(y|x)}{\pi_{\mathrm{ref}}(y|x)}$ \citep{rafailov2024direct}, it shows that the learning objective in token-wise MDP equivalents to a BT model with a token-wise reward function 
\begin{align}
r^*(s_h = (x, y_{1:h-1}), a_h = y_h) = \beta_\text{reg} \log \frac{\pi_{\beta_\text{reg}}^*(a_h|s_h)}{\pi_{\mathrm{ref}}(a_h|s_h)} = \beta_\text{reg} \log \frac{\pi_{\beta_\text{reg}}^*(y_h|x, y_{1:h-1})}{\pi_{\mathrm{ref}}(y_h|x, y_{1:h-1})},
\end{align}
where $x$ is the prompt, $y_{1:h-1}$ is the tokens generated so far, and $y_h$ is the token chosen at the current step. RTO assigns the defined token-wise reward function to each step. Formally, for any $h$, it is defined as following:
\begin{equation}
    \begin{aligned}\label{eq:prac:5}
&{\beta_\text{reg}}^1 \log \frac{\pi_{\beta_\text{reg}}^*(y_h|x, y_{1:h-1})}{\pi_{\mathrm{ref}}(y_h|x, y_{1:h-1})} - \beta_{\text{reg}}^2 \log \frac{\pi(y_h|x, y_{1:h-1})}{\pi_{\mathrm{ref}}(y_{h}|x, y_{1:h-1})}  \\
& \qquad \approx \beta_\text{reg}^1 \log \frac{\pi_{\mathrm{dpo}}(y_h|x, y_{1:h-1})}{\pi_{\mathrm{ref}}(y_h|x, y_{1:h-1})} - \beta_{\text{reg}}^2 \log \frac{\pi(y_h|x, y_{1:h-1})}{\pi_{\mathrm{ref}}(y_{h}|x, y_{1:h-1})} := r_{\mathrm{rto}}((x, y_{1:h-1}),y_h),
    \end{aligned}
\end{equation}
as the token-wise reward used by RTO,
where $\beta_\text{reg}^1$ and $\beta_{\text{reg}}^2$ are tuning parameters, and $\pi$ is the current policy to be updated. In the last step, RTO uses $\pi_{\mathrm{dpo}}$, the policy learned by DPO, as a proxy for the unknown optimal $\pi_{\beta_\text{reg}}^*$. Then RTO employs PPO to optimize the model with respect to the token-wise reward $r_{\mathrm{rto}}$. The idea of transformation from sequence level preferences to token level guidance also appeared in an earlier work by~\cite{yang2024preference}.

\section{Evaluation}
Evaluation metrics and pipelines are essential for measuring the core capabilities of LLMs in performing tasks and assessing their alignment with human preferences in open-ended scenarios. Numerous evaluation metrics have been proposed in the literature. In this section, we will describe these metrics and evaluation methods across different modalities.

\subsection{LLM As A Judge}
Human evaluation is both costly and time-consuming. Developing an automatic evaluation method that closely aligns with human assessments can significantly reduce evaluation time and accelerate research progress. In this context, we outline the benchmarks employed for automatic evaluation using LLMs.

\subsubsection{AlpacaEval}
AlpacaEval~\citep{dubois2024length} win rate (against GPT4) is an LLM-based automatic evaluation that has high-level agreement to human. To further improve the fairness of the evaluation and address the verbosity of issue of GPT4 as a judge, \cite{dubois2024length} introduce a length-controlled version of AlpacaEval that aims to conduct measurement with outputs with similar lengths. The metric is used in AlpacaEval calculates win-rates for models across a variety of NLP tasks to measure of model capabilities compared to a baseline by using an LLM judge. \textbf{AlpacaEval 2.0:} The judge uses GPT4-Turbo to replace GPT-3 based model ``text-davinci-003" in the 1.0 version, which makes it more challenging and have a metric that better reflects the current SOTA model.

\subsubsection{ChatbotArena}

ChatbotArena~\citep{chiang2024chatbot} is a benchmarking platform for Large Language Models (LLMs) that conducts anonymous, randomized battles in a crowdsourced environment. On this platform, users can pose questions and receive responses from two anonymous LLMs. After reviewing the answers, users vote for the response they prefer, with the identities of the models revealed only after voting. This crowdsourced approach effectively gathers a diverse array of user prompts, accurately reflecting real-world LLM applications. Utilizing this data, they apply a range of advanced statistical techniques, from the Bradley-Terry model~\citep{bradley1952rank} to the E-values framework~\citep{vovk2021values}, to estimate model rankings as reliably and efficiently as possible.

\subsubsection{MT-Bench}
MT-bench~\citep{zheng2024judging} is a series of open-ended questions designed to evaluate a chatbot’s multi-turn conversational and instruction-following abilities. It is used in the platform that assesses these capabilities in a crowdsourced battle format. This platform is particularly useful for evaluating the quality of LLM-generated responses, utilizing judges like GPT-4. Consequently, employing LLM as a judge provides a scalable and explainable method to approximate human preferences, which would otherwise be very costly to obtain.

\subsubsection{HELM}
HELM~\citep{liang2022holistic} is a large-scale reproducible and transparent framework for evaluating LLM models to enhance the transparency of language models. The framework has seven metrics, such as accuracy, calibration, robustness, fairness, bias, toxicity, and efficiency.

\subsection{Vision Language Model Evaluation}
\subsubsection{VHELM}
VHELM\footnote{\url{https://crfm.stanford.edu/helm/vhelm/latest/}.} is an extension of the HELM framework~\citep{liang2022holistic} with the adaptation methods to assess the performance of VLMs by scoring the winning rates against the GPT-4V model.

\subsubsection{MMStar}
MMStar~\citep{chen2024we} is a multi-modal benchmark consisting of 1,500 samples meticulously curated by human experts. It evaluates six core capabilities and 18 specific criteria to assess the multi-modal capacities of LVLMs. The samples are selected from existing benchmarks using an automated process, followed by human review to ensure each sample demonstrates visual dependency, minimal data leakage, and requires advanced multi-modal skills.

\subsection{Speech Language Model Evaluation}
\subsubsection{SpeechLMScore}
SpeechLMScore~\citep{maiti2023speechlmscore} calculates the average log-probability of a speech signal by converting it into discrete tokens and assessing the average probability of generating the token sequence. Formally, SpeechLMScore($\mathbf{x}| \theta$) is defined as: 
\begin{equation} \text{SpeechLMScore}(\mathbf{d}|\theta) =\frac{1}{T} \sum_{i=1}^{T} \log p(d_i | d_{<i}, \theta), 
\end{equation} 
where $\theta$ is an LM used to generate the score. Specifically, to compute SpeechLMScore, the process involves: i) encoding the speech into discrete tokens $\mathbf{d} = {d_1 \cdots d_T}$, and ii) iteratively calculating the log probability of each token $d_{i}$ given all preceding tokens ${d_1 \cdots d_{i-1}}$ using $\theta$, i.e., $\log p(d_i | d_{<i}, \theta)$. SpeechLMScore thus measures the average log-probability of a sequence of speech tokens. This metric is closely related to the perplexity of a speech sample, essentially indicating how perplexed a speech language model is when presented with a set of discrete tokens from speech $\mathbf{x}$.

\subsubsection{SpeechBERTScore}
SpeechBERTScore~\citep{saeki2024speechbertscore} evaluates the BERTScore for self-supervised dense speech features derived from both generated and reference speech, even when these sequences differ in length. This method utilizes BERTScore as a metric to assess the quality of speech generation. By computing the BERTScore for SSL feature sequences from both the generated and reference speech, SpeechBERTScore effectively captures their semantic alignment.

\subsection{Reward Model Evaluation}
One way to assess the quality of our model is by evaluating the performance of the reward model using a benchmark. \cite{zhustarling, jiang2023llm} propose using validation sets from previous RLHF training processes, such as Anthropic’s Helpful and Harmless data~\citep{bai2022training} or OpenAI’s Learning to Summarize~\citep{stiennon2020learning}. Additionally, newly released preference data, aimed at expanding the diversity of preference training datasets, such as UltraFeedback~\citep{cui2023ultrafeedback}, UltraInteract~\citep{yuan2024advancing}, and Nectar~\citep{zhustarling}, lack test sets, necessitating a new style of evaluation for reward models. RewardBench is a benchmark dataset and codebase designed for this purpose~\citep{lambert2024rewardbench}. The dataset comprises a collection of prompt-chosen-rejected triplets that span various domains, including chat, reasoning, and safety. This allows for a comprehensive evaluation of how reward models perform on challenging, structured, and out-of-distribution queries. \cite{winata2024metametrics} propose \textsc{MetaMetrics}, a new method to construct a meta-metric that is aligned with human preferences by calibrating multiple metrics by using Bayesian optimization and boosting methods, which has been further applied to machine translation~\citep{anugraha2024metametrics}.

\section{Discussion and Research Directions}
In this section, we describe topics related to human preferences that are either underexplored or still in their early stages. We also discuss potential future research areas that could be highly beneficial for advancing the field.

\subsection{Discussion}

\subsubsection{Effectiveness of Optimization Components}
In the literature on preference tuning, the comparative performance of different methods remains unclear, particularly when comparisons are not conducted under fair conditions. This is largely because RL is highly sensitive to changes in hyper-parameters, and running multiple hyper-parameter configurations is very costly. For instance, when a new method is proposed, the baseline may not be fully optimized, resulting in weaker baselines. Another issue in automatic evaluation using LLMs as judges is the bias introduced by the pre-training data. A model might prefer predictions generated by a similar type of model. For example, a GPT-4 model may favor outputs from its own model family over those from other models, such as Llama. Additionally, judge models may have a preference for longer sequences or text in certain positions~\citep{zheng2024judging}. Therefore, finding a less biased model is crucial during evaluation. Consequently, the effectiveness of each method, along with their optimized components and the models used in automatic evaluation, needs further investigation and careful consideration.

\subsubsection{Offline vs. Online Algorithms}
Through theoretical and experimental analysis, \cite{xu2024dpo} explore the limitations of DPO and find that DPO is sensitive to distribution shifts between the base model outputs and preference data. They suggest that iterative DPO, which involves continuous updating, is more effective than training on static data. However, they also find that DPO fails to improve performance on challenging tasks such as code generation. From a different perspective, \cite{tang2024understanding} clarify the confusion surrounding the limitations of offline algorithms' performance, often attributed to the bounded performance of offline algorithms. The paper discusses that the dichotomy between online and offline algorithms is frequently inaccurate in practice. An offline algorithm that continuously updates its data stream effectively functions as an online algorithm. Consequently, the shortcomings identified in offline learning can be mitigated by adopting a more careful approach to the data generation process.

\subsection{Research Directions}
Here, we explore potential research directions that offer significant opportunities for further investigation and development. These avenues hold promise for both academic researchers and industry practitioners, providing ground for innovative studies and practical applications. We summarize key ideas and topics that could drive future advancements in the field, highlighting areas where there is ample room for exploration and growth.

\subsubsection{Multilingual, Multicultural, and Pluralistic Preference Tuning}
While significant resources have been allocated to enhance the safety of LLMs for deployment, the safety of multilingual LLMs remains underexplored. \cite{ahmadian2024multilingual} is one of the pioneering works pushing the boundaries of aligning language models by optimizing for both general and safety performance simultaneously in a multilingual setting using Distributional DPO. Similarly, \cite{li2024preference} propose exploring DPO training to reduce toxicity in multilingual open-ended generations. Another line of research focuses on using multilingual alignment based on human preferences to improve reasoning abilities, aiming to align reasoning processes in other languages with those in the dominant language~\citep{she2024mapo}. There is still ample room for exploration in the multilingual space, particularly in examining the cultural aspects of multilingualism~\citep{adilazuarda2024towards,alkhamissi2024investigating} and improving the alignment of LLM for generation~\citep{winata2021language}. It is crucial to cover more diverse languages, including regional languages, different dialects~\citep{aji2022one}, and code-switching~\citep{winata2021multilingual}, which are common phenomena in bilingual and multilingual communities~\citep{winata2024miners}. Additionally, the exploration of multilingual topics in vision-language and speech tasks remains open for further investigation.

\subsubsection{Multi-modality}
While alignment in LLMs has been extensively studied, alignment for multi-modal models has not yet been investigated to the same extent. \cite{sun2023aligning} and \cite{zhou2024aligning} align LLaVA \citep{liu2024improved} using PPO and DPO, respectively. Similarly, \cite{li2023silkie} and \cite{yu2023reformulating} employ DPO and its variations to align the Qwen-VL~\citep{bai2023qwen} and Muffin~\citep{yu2023reformulating} models. Notably, in addition to different alignment strategies and base models, all these works introduce novel preference datasets for alignment, varying in size, collection methods, and generation schemes. Consequently, while each of these studies offers valuable insights into alignment for multi-modal LLMs, it can sometimes be challenging to attribute reported improvements to specific proposed choices. Furthermore, \cite{amirloo2024understanding} examine each component of multi-modal alignment independently, involving sampling from the model during policy optimization.

\subsubsection{Speech Applications}
The application of preference tuning in speech technology is in its early stages, offering many opportunities for future exploration. As research advances, preference tuning is expected to enhance various speech-related technologies, including TTS and speech recognition systems, by incorporating human preferences to improve performance and user satisfaction. In TTS, it can help select the most natural and pleasing synthetic voices~\citep{zhang2024speechalign}, while in speech recognition, it can ensure more accurate and contextually appropriate transcriptions. Additionally, preference tuning can benefit voice assistants, automated customer service systems, and language learning tools by creating more intuitive and effective interfaces. Ongoing research and experimentation will be essential to fully realize the potential of preference tuning in speech technology, aiming to develop systems that are both technically proficient and closely aligned with human communication and preferences.

\subsubsection{Unlearning}
\cite{yao2023large,zhang2024negative} propose an alignment technique for unlearning by utilizing negative examples, which are easier and cheaper to collect than the positive examples needed for preference tuning. This method is considered computationally efficient, with costs comparable to light supervised finetuning. They demonstrate that unlearning is particularly appealing when resources are limited and the priority is to stop generating undesirable outputs. Despite using only negative samples, unlearning can achieve better alignment performance than RLHF. The unlearning method can be very useful in removing harmful responses, erasing copyright-protected content, and reducing hallucinations. This approach is promising and has potential for further exploration in future work.

\subsubsection{Benchmarking Preference Tuning Methods}
Developing a comprehensive benchmark for various preference tuning methods is essential for gaining a clearer understanding of their individual effectiveness. Currently, the effectiveness of each method is somewhat unclear, making it difficult to fully appreciate their value. By creating a benchmark, we can systematically assess and compare these methods, thereby clarifying their strengths and weaknesses. This effort to elucidate the usefulness of each approach is vital for advancing our knowledge and improving the application of preference tuning techniques. Such a benchmark would not only enable more informed decisions when selecting the most suitable method for specific tasks but also stimulate innovation by identifying areas that require further refinement and development. Ultimately, this initiative aims to enhance the overall effectiveness and reliability of preference tuning methods across various applications.

\subsubsection{Mechanistic Understanding of Preference Tuning Methods}
Despite the popularity of preference tuning methods for LLM alignment, explanations for their underlying mechanisms in which models become ``aligned" still lack, thus making it difficult to explain phenomena like jailbreaks \citep{chao2023jailbreaking}. Taking toxicity reduction as the task and applying DPO on GPT2-medium, \cite{lee2024mechanistic} suggest that capabilities may be rather bypassed instead of removed. Thus, current preference tuning methods may still be vulnerable to reverse-engineering and the models tuned are easy to be unaligned again. More interpretation of preference tuning methods could possibly address these concerns by ensuring that models not only meet alignment objectives more reliably but also provide clearer insights into how these objectives are achieved; it could also possibly help lead to better preference tuning methods that can mitigate issues such as jailbreaking and other forms of misalignment, where models exhibit undesirable behaviors despite appearing aligned during training.

\bigskip
\noindent
{\bf Acknowledgments}: 
Wenpin Tang and Hanyang Zhao are supported by NSF grant DMS-2206038, a start-up grant at Columbia University, and the Columbia Innovation Hub grant. The works of Hanyang Zhao and David D. Yao are part of a Columbia-CityU/HK collaborative project that is supported by InnotHK Initiative, The Government of the HKSAR and the AIFT Lab. 

\newpage

\bibliography{main}

\end{document}